\RequirePackage[hyphens]{url}

\documentclass[sw]{arxiv}

\usepackage[utf8]{inputenc}
\usepackage{amssymb}
\usepackage{tabularx}
\usepackage{longtable}
\usepackage{adjustbox}
\usepackage{amsmath}
\usepackage{parskip}
\usepackage{booktabs}
\usepackage[many]{tcolorbox}
\usepackage{cleveref}
\usepackage{float}
\floatstyle{plaintop}
\restylefloat{table}
\usepackage{subcaption}
\usepackage{multicol}
\usepackage{adjustbox}
\usepackage{diagbox}
\usepackage[multiple]{footmisc}
\usepackage[flushleft, referable]{threeparttablex}
\usepackage[htt]{hyphenat}
\usepackage[section]{placeins}
\usepackage{xcolor}
\usepackage{stfloats}
\usepackage{suffix}
\usepackage{pifont}
\usepackage{sistyle}
\SIthousandsep{,}
\usepackage{array}
\newcolumntype{P}[1]{>{\centering\arraybackslash}p{#1}}

\newcommand{\cmark}{\ding{51}}%
\newcommand{\xmark}{\ding{55}}%
\newcommand{\citea}[1]{\citeauthor{#1}~(\citeyear{#1})~\cite{#1}}
\WithSuffix\newcommand\citea*[1]{\citeauthor{#1}~\cite{#1}}
\newcolumntype{Y}{>{\centering\arraybackslash}X}

\DeclareMathOperator*{\argmax}{arg\,max}

\newtcolorbox{highlightboxm}[1][]{
    enhanced,
    title={#1},
    coltitle=black,
    boxrule=0pt,
    fonttitle=\bfseries,
    breakable,
    top=0pt,
    boxsep=0pt,
    topsep at break=3pt,
    left=2pt,
    right=2pt,
    colbacktitle=green!40!brown!30,
    colback=green!20!brown!10,
    toptitle=2pt,
    bottomtitle=2pt,
    arc=1mm,
    sharp corners=south,
    colframe=white,
    attach boxed title to top center={xshift=-0.00mm, yshift=-0.8mm,}, 
    boxed title style={sharp corners=south,
    },
    subtitle style={
        colback=green!40!brown!30}
}

\newenvironment{highlightbox}[2]
    {
        \begin{highlightboxm}[{#1}]
            \tcbsubtitle[boxrule=0pt, toprule=-7pt, colframe=green!40!brown!30, halign=left, arc=1mm,
            fontupper=]{#2}
    }
    {        
        \end{highlightboxm}
    }

\pubyear{2021}
\volume{0}
\firstpage{1}
\lastpage{1}

\begin{document}
\begin{frontmatter}
{\color{blue}Disclaimer: © Cedric Möller, Jens Lehmann, Ricardo Usbeck, 2021. The definitive, peer reviewed and edited version of this article is published in the Semantic Web Journal, Special issue: Latest Advancements in Linguistic Linked Data, 2021}

\title{Survey on English Entity Linking on Wikidata}
\runtitle{Survey on English Entity Linking on Wikidata}
\subtitle{Datasets and Approaches}

\author[A]{\inits{C.}\fnms{Cedric} \snm{Möller}\ead[label=e1]{cedric.moeller@uni-hamburg.de}%
\thanks{Corresponding author. \printead{e1}.}},
\author[B,C]{\inits{N.N.}\fnms{Jens} \snm{Lehmann}\ead[label=e2]{jens.lehmann@cs.uni-bonn.de}\ead[label=e4]{jens.lehmann@iais.fraunhofer.de}}
and
\author[A,D]{\inits{N.-N.}\fnms{Ricardo} \snm{Usbeck}\ead[label=e3]{firstname.lastname@uni-hamburg.de}}

\address[A]{Semantic Systems Group, \orgname{Universität Hamburg}, Mittelweg 177, 20148 Hamburg, \cny{Germany} \printead[presep={\\}]{e3}}

\address[B]{NetMedia Department, \orgname{Fraunhofer IAIS}, Zwickauer Stra\ss{}e 46, 01069 Dresden, \cny{Germany} \printead[presep={\\}]{e4}}

\address[C]{\orgname{University of Bonn}, Endenicher Allee 19a, 53115, Bonn, \cny{Germany} \printead[presep={\\}]{e2}}

\address[D]{\orgname{HITeC Hamburg e.V.}, Vogt-Kölln-Straße 30, 22527 Hamburg, \cny{Germany}}

\vspace{-5mm}

\begin{review}{editor}
\reviewer{\fnms{First} \snm{Editor}\address{\orgname{University or Company name}, \cny{Country}}}
\reviewer{\fnms{Second} \snm{Editor}\address{\orgname{First University or Company name}, \cny{Country}
    and \orgname{Second University or Company name}, \cny{Country}}}
\end{review}
\begin{review}{solicited}
\reviewer{\fnms{First} \snm{ Reviewer}\address{\orgname{University or Company name}, \cny{Country}}}
\reviewer{\fnms{Second} \snm{ Reviewer}\address{\orgname{University or Company name}, \cny{Country}}}
\reviewer{\fnms{Third } \snm{ Reviewer}\address{\orgname{University or Company name}, \cny{Country}}}
\end{review}
\vspace{-15mm}
\begin{abstract}
Wikidata is a frequently updated, community-driven, and multilingual knowledge graph. Hence, Wikidata is an attractive basis for Entity Linking, which is evident by the recent increase in published papers. This survey focuses on four subjects: (1) Which Wikidata Entity Linking datasets exist, how widely used are they and how are they constructed? (2) Do the characteristics of Wikidata matter for the design of Entity Linking datasets and if so, how?
(3) How do current Entity Linking approaches exploit the specific characteristics of Wikidata? (4) Which Wikidata characteristics are unexploited by existing Entity Linking approaches? 
This survey reveals that current Wikidata-specific Entity Linking datasets do not differ in their annotation scheme from schemes for other knowledge graphs like DBpedia.
Thus, the potential for multilingual and time-dependent datasets, naturally suited for Wikidata, is not lifted.
Furthermore, we show that most Entity Linking approaches use Wikidata in the same way as any other knowledge graph missing the chance to leverage Wikidata-specific characteristics to increase quality. Almost all approaches employ specific properties like labels and sometimes descriptions but ignore characteristics such as the hyper-relational structure. Hence, there is still room for improvement, for example, by including hyper-relational graph embeddings or type information.
Many approaches also include information from Wikipedia, which is easily combinable with Wikidata and provides valuable textual information, which Wikidata lacks.
\end{abstract}

\end{frontmatter}

\section{Introduction}
\subsection{Motivation}
\begin{figure}
    \centering
    \includegraphics[width=\linewidth, trim={0 2cm 0 0},clip]{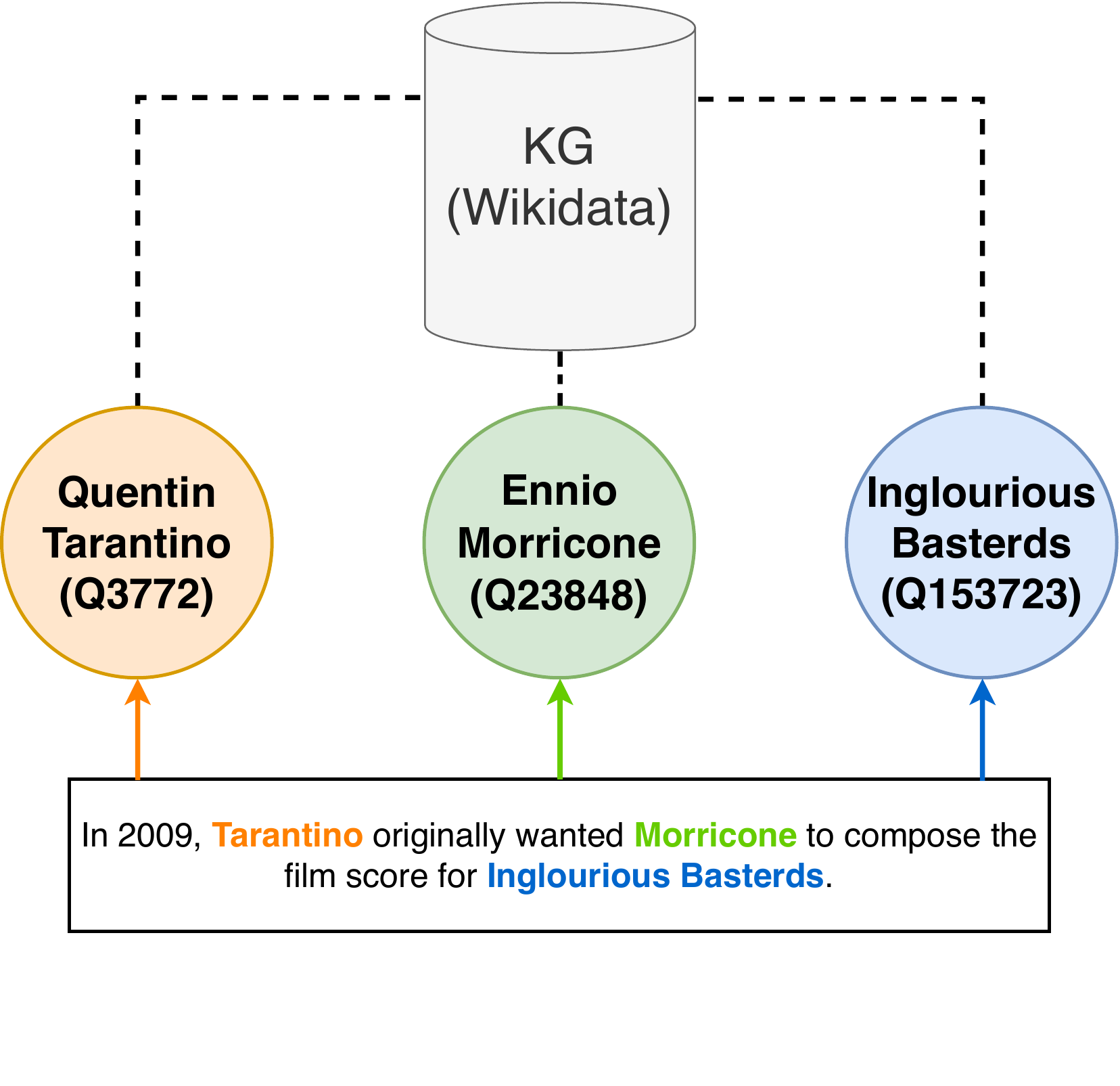}
    \caption{Entity Linking - Mentions in the text are linked to the corresponding entities (color-coded) in a knowledge graph (here: Wikidata).}
    \label{fig:ELExample}
\end{figure}
Entity Linking (EL) is the task of connecting already marked mentions in an utterance to their corresponding entities in a knowledge graph (KG), see Figure~\ref{fig:ELExample}. In the past, this task was tackled by using popular knowledge bases such as DBpedia~\cite{DBLP:journals/semweb/LehmannIJJKMHMK15}, Freebase~\cite{DBLP:conf/sigmod/BollackerEPST08} or Wikipedia. While the popularity of those is still imminent, another alternative, named Wikidata~\cite{DBLP:journals/cacm/VrandecicK14}, appeared. 

\begin{figure}[htb!]
    \centering
    \includegraphics[width=\linewidth]{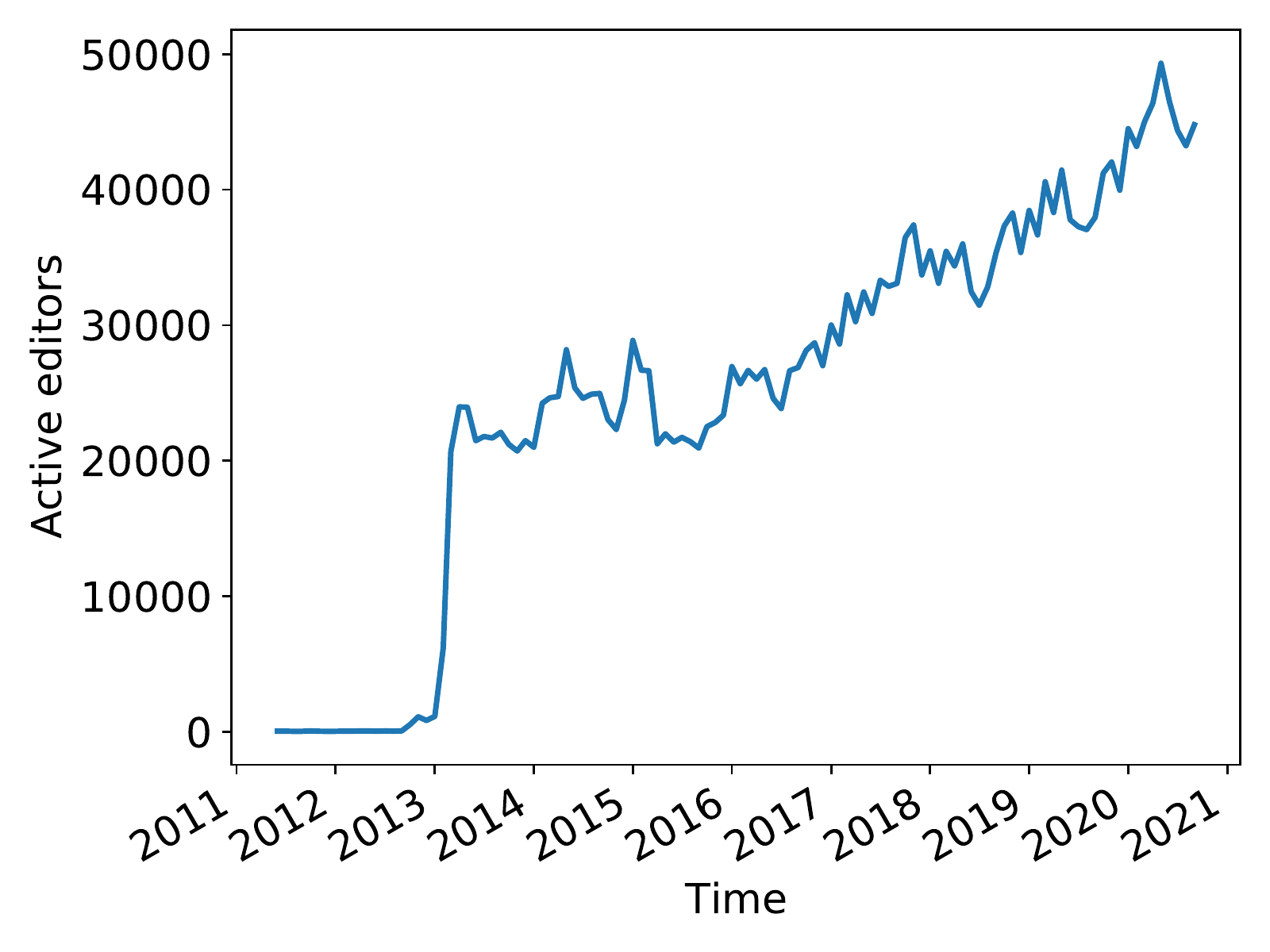}
    \caption{Active editors in Wikidata~\cite{Foundation2020}.}
    \label{fig:editors}
\end{figure}

\begin{figure}[htb!]
    \centering
    \includegraphics[width=\linewidth]{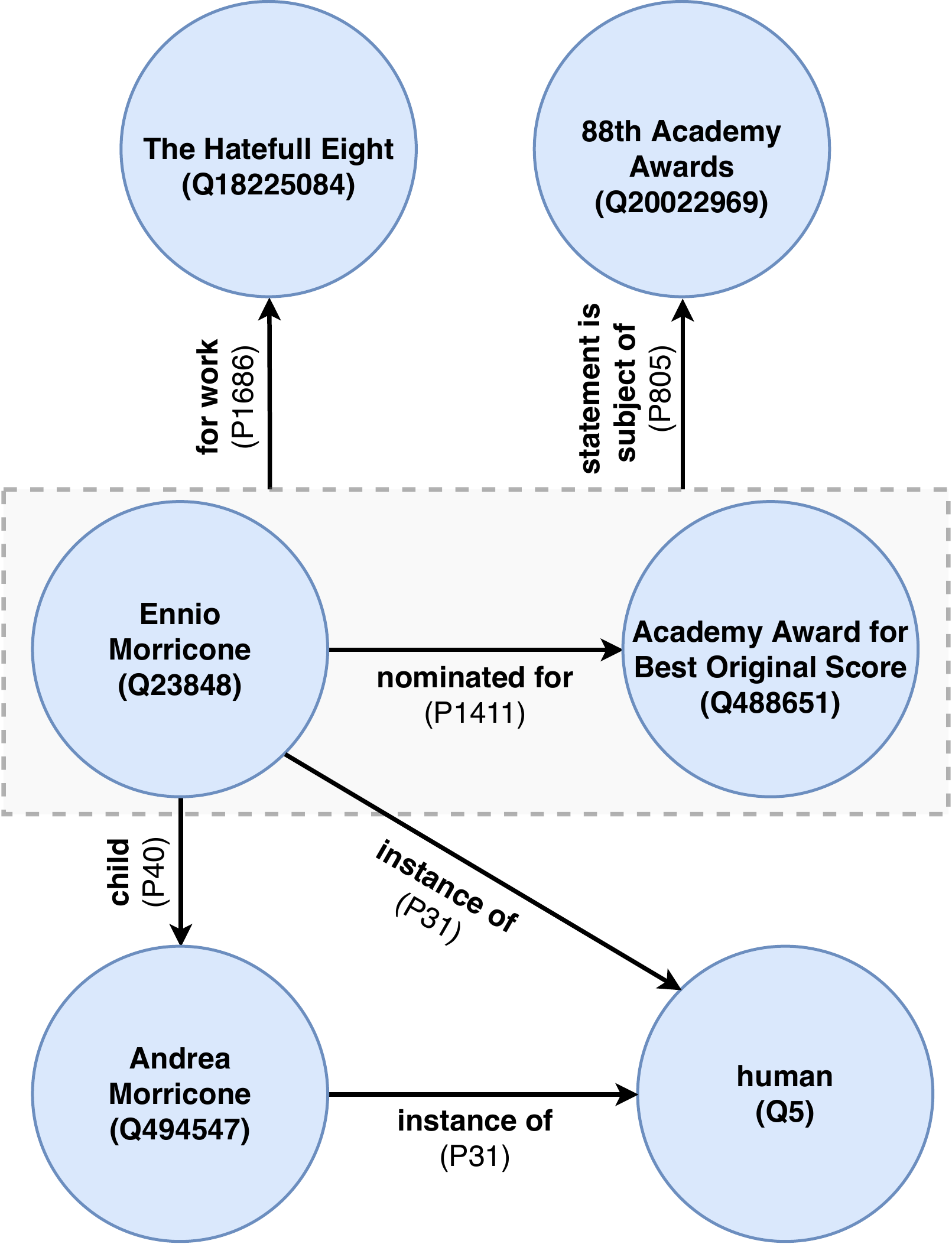}
    \caption{Wikidata subgraph - Dashed rectangle represents a claim with attached qualifiers.}
    \label{fig:wikidata_subgraph}
\end{figure}

Wikidata follows a similar philosophy as Wikipedia as it is curated by a continuously increasing community, see Figure~\ref{fig:editors}. However, Wikidata differs in the way knowledge is stored -  information is stored in a structured format via a knowledge graph (KG). 
An important characteristic of Wikidata is its inherent multilingualism. While Wikipedia articles exist in multiple languages, Wikidata information are stored using language-agnostic identifiers. This is of advantage for multilingual entity linking. 
DBpedia, Freebase or Yago4~\cite{Tanon2020} are KGs too which can become outdated over time~\cite{Ringler2017}. They rely on information extracted from other sources in contrast to the Wikidata knowledge which is inserted by a community. 
Given an active community, this leads to Wikidata being frequently and timely updated - another characteristic.
Note that DBpedia also stays up-to-date but has a delay of a month\footnote{\url{https://release-dashboard.dbpedia.org/}} while Wikidata dumps are updated multiple times a month.
There are up-to-date services to access knowledge for both KGs, Wikidata and DBpedia (cf. DBpedia Live~~\footnote{\url{https://wiki.dbpedia.org/online-access/DBpediaLive}}), but full dumps are preferred as else the FAIR replication~\cite{wilkinson2016fair} of research results based on the KG is hindered. 
Another Wikidata characteristic interesting for Entity Linkers, are hyper-relations (see Figure~\ref{fig:wikidata_subgraph} for an example graph), which might affect their abilities and performance. 

Therefore, it is of interest how existing approaches incorporate these characteristics.
However, existing literature lacks an exhaustive analysis which examines Entity Linking approaches in the context of Wikidata.

Ultimately, this survey strives to expose the benefits and associated challenges which arise from the use of Wikidata as the target KG for EL. 
Additionally, the survey provides a concise overview of existing EL approaches, which is essential to (1) avoid duplicated research in the future and (2) enable a smoother entry into the field of Wikidata EL. 
Similarly, we structure the dataset landscape which helps researchers find the correct dataset for their EL problem.

\begin{figure}[htb!]
    \centering
    \includegraphics[width=\linewidth]{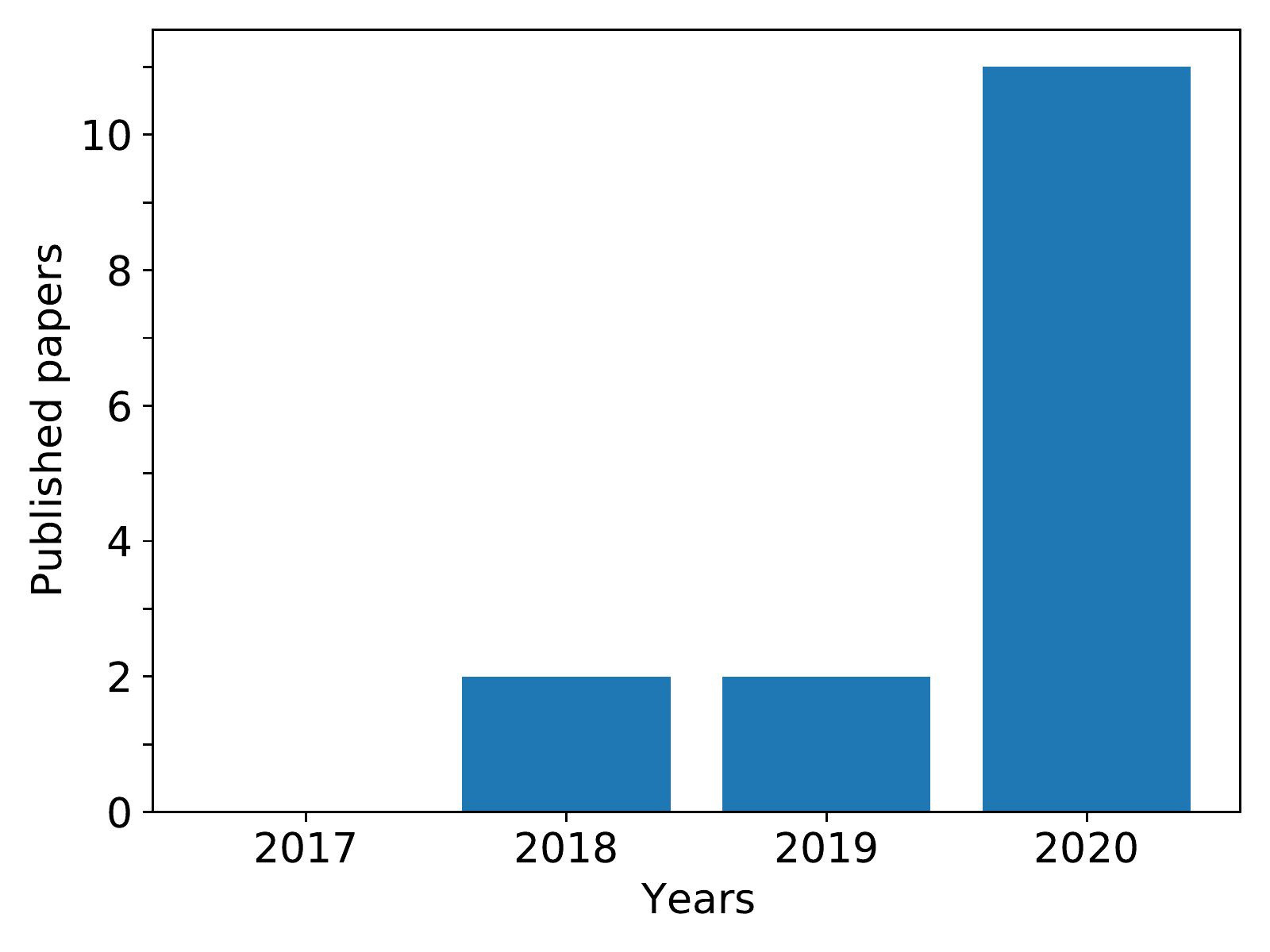}
    \caption{Publishing years of included Wikidata EL papers (Table~\ref{tab:comparison_approaches_wikidata}).}
    \label{fig:papers}
\end{figure}

The focus of this survey lies on EL approaches, which operate on already marked mentions of entities, as the task of Entity Recognition~(ER) is much less dependent on the characteristics of a KG. However, due to the recent uptake of research on EL on Wikidata, there is only a low number of EL-only publications. To broaden the survey's scope, we also consider methods that include the task of ER. We do not restrict ourselves regarding the type of models used by the entity linkers. 

This survey limits itself to all EL approaches supporting the English language as most frequent language, and thus, a better comparison of the approaches and datasets is possible. We also include approaches that support multiple languages. The existence of such approaches for Wikidata is not surprising as an important characteristic of Wikidata is the support of a multitude of languages.

\subsection{Research Questions and Contributions}
First, we want to develop an overview of datasets for EL on Wikidata. Our survey analyses datasets and whether they are designed with Wikidata in mind and if so, in what way?
Thus, we post the following two research questions:
\begin{quote}
    \hypertarget{rq1}{\textbf{RQ 1}: Which Wikidata EL datasets exist, how widely used are they and how are they constructed?} 
\end{quote}
\begin{quote}
    \hypertarget{rq2}{\textbf{RQ 2}: Do the characteristics of Wikidata matter for the design of EL datasets and if so, how?}
\end{quote}
To answer those two research questions, an overview of the structure of Wikidata and the amount of information it contains (see Section~\ref{sec:wikidata}) is given. All current Wikidata-specific EL datasets are gathered and analyzed with the research questions in mind. Furthermore, we discuss how the characteristics of Wikidata might affect the design of datasets (see Section~\ref{sec:datasets}).

EL approaches use many kinds of information like labels, popularity measures, graph structures, and more. This multitude of possible signals raises the question of how the characteristics of Wikidata are used by the current state of the art of EL on Wikidata. Thus, the third research question is:
\begin{quote}
    \hypertarget{rq3}{\textbf{RQ 3}: How do current Entity Linking approaches exploit the specific characteristics of Wikidata?}
\end{quote}
In particular, which Wikidata-specific characteristics contribute to the solution? Wikidata-specific characteristics mean characteristics that are part of Wikidata but not necessarily only occurring in Wikidata.

Lastly, we identify what kind of characteristics of Wikidata are of importance for EL but are insufficiently considered. 
This raises the last research question:
\begin{quote}
    \hypertarget{rq4}{\textbf{RQ 4}: Which Wikidata characteristics are unexploited by existing Entity Linking approaches?}
\end{quote}
The last two questions are answered by gathering all existing approaches working on Wikidata systematically, analyzing them, and discussing the potential and challenges of Wikidata for EL (see Section~\ref{sec:approaches}).

This survey makes the following contributions:
\begin{itemize}
    \item An overview of all currently available EL datasets focusing on Wikidata
    \item An overview of all currently available EL approaches linking on Wikidata 
    \item An analysis of the approaches and datasets with a focus on Wikidata characteristics
    \item A concise list of future research avenues
\end{itemize}

\section{Survey Methodology}
There exist several different ways in which a survey can contribute to the research field~\cite{Kitchenham2004}:
\begin{enumerate}
    \item \label{cont_1} Providing an overview of current prominent areas of research in a field
    \item \label{cont_2} Identification of open problems
    \item \label{cont_3} Providing a novel approach tackling the extracted open problems (in combination with the identification of open problems)
\end{enumerate}
We analyse different recent and older surveys on EL and highlight specific areas which are not covered as well as our survey's novelties (see also Section~\ref{sec:discussion}). While some very recent surveys exist~\cite{al2020named, oliveira2020towards, DBLP:journals/corr/abs-2006-00575}, they do not consider the different underlying Knowledge Graphs as a significant factor affecting the performance of EL approaches. Furthermore, barely any approaches included in other surveys are working on Wikidata and take the particular characteristics of Wikidata into account (see Section~\ref{sec:related-work}). 
Our survey fills these gaps by contributing according to \Cref{cont_1,cont_2}.

\begin{table*}[htb!]
    \centering
    \begin{tabularx}{.78\textwidth}{Y Y}
        \toprule
        \multicolumn{2}{c}{\textbf{Criteria}} \\ \midrule
         \textbf{Must satisfy all} & \textbf{Must not satisfy any} \\ 
         \midrule
         \begin{itemize}
             \item Approaches that consider the problem of unstructured EL over Knowledge Graphs 
             \item Approaches where the target Knowledge Graph is Wikidata 
         \end{itemize}
         &
         \begin{itemize}
             \item Approaches conducting Semi-structured EL
             \item Approaches not doing EL in the English language
         \end{itemize} \\
         \bottomrule
    \end{tabularx}
    \caption{Qualifying and disqualifying criteria for approaches. "Semi-structured" in this table means that the entity mentions do not occur in natural language utterances but in more structured documents such as tables.}
    \label{tab:qual_disqual}
\end{table*}

\begin{table*}[htb!]
    \centering
    \begin{tabularx}{.78\textwidth}{YY}
        \toprule
        \multicolumn{2}{c}{\textbf{Criteria}} \\ \midrule
         \textbf{Must satisfy all} & \textbf{Must not satisfy any} \\ \midrule
         \begin{itemize}
             \item Datasets that are designed for EL or are used for evaluation of Wikidata EL
             \item Datasets must include Wikidata identifiers from the start; an existing dataset later mapped to Wikidata is not permitted
         \end{itemize} &
         \begin{itemize}
             \item Datasets without English utterances
         \end{itemize} \\
         \bottomrule
    \end{tabularx}
    \caption{Qualifying and disqualifying criteria for the dataset search.}
    \label{tab:qual_disqual_datasets}
\end{table*}

Until December 18, 2020, we continuously searched for existing and newly released scientific work suitable for the survey.
Note, this survey includes only scientific articles that were accessible to the authors.\footnote{\url{https://www.projekt-deal.de/about-deal/}}

\subsection{Approaches}
Our selection of approaches stems from a search over the following search engines:
\begin{itemize}
    \item Google Scholar
    \item Springer Link
    \item Science Direct 
    \item IEEE Xplore Digital Library
    \item ACM Digital Library
\end{itemize}

To gather a wide choice of approaches, the following steps were applied.
\texttt{Entity Linking}, \texttt{Entity Disambiguation} or variations of the phrases~\footnote{Google Scholar search query: \texttt{(intitle:"entity" OR intitle:"entities") AND (intitle:"link" OR intitle:"linking" OR intitle:"disambiguate" OR entitle:"disambiguation") AND intext:"Wikidata"}} had to occur in the title of the paper. The publishing year was not a criterion due to the small number of valid papers and the relatively recent existence of Wikidata. Any approach where \texttt{Wikidata} was not occurring once in the full text was not considered.  
The systematic search process resulted in exactly 150 papers and theses (including duplicates). 

Following this search, the resulting papers were filtered again using the qualifying and disqualifying criteria which can be found in Table~\ref{tab:qual_disqual}. This resulted in 15 papers and one master thesis in the end.
 
The search resulted in papers in the period from 2018 to 2020. While there exist EL approaches from 2016~\cite{spitz2016so, almeida2016streets} working on Wikidata, they did not qualify according to the criteria above. 

\subsection{Datasets}

The dataset search was conducted in two ways. 
First, a search for potential datasets was performed via the same search engines as used for the approaches.
Second, all datasets occurring in the system papers were considered if they fulfilled the criteria.
The criteria for the inclusion of a dataset can be found in Table~\ref{tab:qual_disqual_datasets}. 

We filtered the dataset papers in the following way. First, in the title, \texttt{Entity Linking} or \texttt{Entity Disambiguation} or variations thereof had to occur, similar to the search for the Entity Linking approaches. Additionally, \texttt{dataset}, \texttt{data}, \texttt{corpus} or \texttt{benchmark} had to occur once in title~\footnote{Google Scholar Search Query: \texttt{intext:"Wikidata" AND (intitle:dataset OR intitle:data OR intitle:benchmark OR intitle:corpus) AND (intitle:entity OR intitle:entities) AND (intitle:link OR intitle:linking OR intitle:disambiguate OR intitle:disambiguation)}} must occur in the title and \texttt{Wikidata} has to appear at least once in the full text. Due to those keywords, other datasets suitable for EL, but constructed for a different purpose like KG population, were not included. This resulted in 26 papers (including duplicates). Of those, only two included Wikidata identifiers and focused on English.

Eighteen datasets were accompanying the different approaches. Many of those did not include Wikidata identifiers from the start. This made them less optimal for the examination of the influence of Wikidata on the design of datasets. They were included in the section about the approaches but not in the section about the Wikidata datasets.

After the removal of duplicates, 11 Wikidata datasets were included in the end.

\section{Problem Definition}
\label{sec:problem}
EL is the task of linking an entity mention in unstructured or semi-structured data to the correct entity in a KG. The focus of this survey lies in unstructured data, namely, natural language utterances.

\subsection{General terms}
\paragraph{Utterance.}
An utterance $u$ is defined as a sequence of $n$ words $w$.
\begin{equation*}
    u = (w_0, w_1, ... w_{n-1})
\end{equation*}

\paragraph{Entity.}
There exists no universally agreed-on definition of an entity in the context of EL~\cite{rosales2020fine}. 
According to the Oxford Dictionary, an entity is:
\begin{quote}
    "something that exists separately from other things and has its own identity"~\cite{oxford_entity}
\end{quote}
What elements of a KG correspond to entities depends on the KG itself. 
In the case of Wikidata, we define it as follows:
\begin{quote}
    Any Wikidata item is an entity. 
\end{quote}
In Section~\ref{sec:wikidata}, we further define Wikidata items.
Many EL approaches limit the space of valid entities. Usually, named entities like a specific person (e.g. \texttt{Barack Obama}), an organization (e.g. \texttt{NASA}) or a movie (e.g. \texttt{The Hateful Eight}) are desirable to link. In general, any entity which can be denoted with a proper noun is a named entity. But sometimes, also common entities like concepts (e.g. \texttt{dog} or \texttt{theater}) are included. What exactly is linked depends on the use case~\cite{rosales2020fine}.

\paragraph{Knowledge Graph.}
While the term knowledge graph was already used before, the popularity increased drastically after Google introduced the \texttt{Knowledge Graph} in 2012~\cite{singhal2012introducing, Ehrlinger2016}.  
However, similar to an entity, there exists no unanimous definition of a KG~\cite{Ehrlinger2016, hogan2020knowledge}.  
For example, Färber et al. define a KG as an RDF graph~\cite{DBLP:journals/semweb/FarberBMR18}.
However, a KG being an RDF graph is a strict assumption. While the Wikidata graph is available in the RDF format, the main output format is JSON. Freebase, often called a KG, did not provide the RDF format until a year after its launch~\cite{freebase_rdf}. 
Paulheim defines it less formal as: 
\begin{quote}
    "A knowledge graph (i) mainly describes real world entities and their interrelations, organized in a graph, (ii) defines possible classes and relations of entities in a schema, (iii) allows for potentially interrelating arbitrary entities with each other and (iv) covers various topical domains."~\cite{Paulheim2017}
\end{quote}
But constraint (iv) of Paulheims definition alienates commercial KGs, focusing on a single domain, such as a financial, medical, or geographical one. 
As no unanimously agreed definition exists, we define a knowledge graph very broadly in the following and the Wikidata KG more concrete in Section~\ref{sec:wikidata}.
The term KG is often used as a synonym for the term knowledge base~(KB), but they are not the same~\cite{jarke1989kbms}. No single definition for a KB exists either. Jarke et al. define it as "a representation of heuristic and factual information, often in the form of facts, assertions and deduction rules"~\cite{jarke1989kbms}. The term is often loosely used to describe a system that is able to store knowledge in the form of structured or unstructured information. While any KG is a KB, not any KB is a KG. The main difference is that a KB does not have to be graph-structured.

In this survey, a knowledge graph is defined as a directed graph $G=(V,E, \mathcal{R})$ consisting of vertices $V$, edges $E$ and relations $\mathcal{R}$. A subset of the vertices corresponds to entities $\mathcal{E}$ or literals $\mathcal{L}$. A literal is a concrete value of information like the height or a name of an entity. A literal vertex has incoming edges but no outgoing ones. Other types of vertices might exist depending on the KG.
$E$ is a set $\{e_1, \dots, e_{|E|}\}$ of edges with $e_j \in V \times \mathcal{R} \times V$ where relations $\mathcal{R}$ assign a certain meaning to the connection between entities. Such edges are also called triples. 
There are special subtypes of KGs, e.g., hyper-relational graphs such as Wikidata.

\paragraph{Hyper-Relational Knowledge Graphs. \label{par:hyper}} In a hyper-relational knowledge graph,  statements can be specified by more information than a single relation. Multiple relations are, therefore, part of a statement.
In case of a hyper-relatio\-nal graph $\mathcal{G} = (V, E, \mathcal{R})$, $E$ is a list $(e_1, \dots, e_n)$ of edges with $e_j \in V \times \mathcal{R} \times V \times \mathcal{P}(\mathcal{R} \times V)$ for $1\le j \le n$, where $\mathcal{P}$ denotes the power set. A hyper-relational fact $e_j \in E$ is usually written as a tuple $ (s,r,o,\mathcal{Q})$, where $\mathcal{Q}$ is the set of \emph{qualifier pairs} $\{(qr_{i},qv_{i}) \}$ with \emph{qualifier relations} $qr_{i} \in \mathcal{R}$ and \emph{qualifier values} $qv_{i} \in V$. The triple $(s,r,o)$ is referred to as the \emph{main triple} of the fact. $Q_j$ denotes the qualifier pairs  
of $e_j$~\cite{Galkin2020}. 
For example, the \texttt{nominated for}~edge in Fig.~\ref{fig:wikidata_subgraph} has two additional qualifier relations and would be represented as \nohyphens{\texttt{(Ennio Morricone, nominated for,  Academy Award for Best Original Score, \{( for work, The Hateful Eight ), (statement is subject of, 88th A-cademy Awards)\})}}. 

\subsection{Tasks}
Since not only approaches that solely do EL were included in the survey, Entity Recognition will also be defined. 

\paragraph{Entity Recognition.} ER is the task of identifying the mention span $$m = (w_i, ..., w_k) | 0 \leq i \leq k \leq n-1$$ of all entities in an utterance $u$. Each such span is called an entity mention $m$. The word or word sequence referring to an entity is also known as the surface form of an entity.
 An utterance can contain more than one entity, often also consisting of more than one word. Sometimes, a broad type of an entity is classified too. Usually, those are \texttt{person}, \texttt{location} and \texttt{organization}. Some of the considered approaches do such a classification task and also use it to improve the EL.
 
 It is also up to debate what an entity mention is. In general, a literal reference to an entity is considered a mention. But whether to include pronouns or how to handle overlapping mentions depends on the use case. 

\paragraph{Entity Linking.} 
The goal of EL is to find a mapping function that maps all found mentions to the correct KG entities and also to identify if an entity mention does not exist in the KG. 

In general, EL takes the utterance $u$ and all $k$ identified entity mentions $M=(m_1, ... m_k)$ in the utterance and links each of them to an element of the set $(\mathcal{E}\cup \{\mathit{NIL}\})$. The $\mathit{NIL}$ element is added to the set of vertices to be able to signalize that the entity, that the mention is referring to, is not known to the KG. Such a $\mathit{NIL}$ entity is also called an out-of-KG entity. Another way to handle such unknown entities is to create emerging entities~\cite{DBLP:conf/www/HoffartAW14}. In that case, the entity is still unknown to the KG, but after encountering it, it is separately stored using information like the provided entity mentions. Now no single $\mathit{NIL}$ entity, but a growing set of emerging entities exists. EL is then done using the entities in the KG and all already encountered emerging entities. While all KG-unknown entities point to the same single $\mathit{NIL}$ entity, they might point to different emerging entities.

EL is often split into two subtasks. First, potential candidates for an entity are retrieved from a KG. This is necessary as doing EL over the whole set of entities is often intractable. This \emph{Candidate generation} is usually performed via efficient metrics measuring the similarities between mentions in the utterance and entities in the KG. The result is a set of candidates $C=\{c_0, \cdots, c_l\}$ for each entity mention $m$ in the utterance. 
After limiting the space of possible entities, one of the available candidates is chosen for each entity. This is done via a \emph{candidate ranking} algorithm, which assigns a rank to each candidate.
The assignment is done by computing a score for each candidate signalizing how likely it is the correct entity. The candidate with the highest score is chosen as the correct entity for the mention.

There are two different categories of reranking methods are called \emph{local} or \emph{global}~\cite{DBLP:conf/acl/RatinovRDA11}.  
\begin{align*}
 \mathit{score_{\mathit{local}}}: C \times M &\to \mathbb{R}\\ 
 \text{ given by } (c,m) &\mapsto \mathit{score_{\mathit{local}}}(c, m)
\end{align*}
where $\mathit{score_{\mathit{local}}}$ is a local scoring function of a candidate. The goal is then to optimize the objective function:
\begin{equation*}
    A^* = \argmax_A \sum_{i=1}^k \mathit{score_{\mathit{local}}}(a_i, m_i) | a_i \in C_i
\end{equation*}
where $A = \{a_1, ... , a_k\} \in \mathcal{P}(\mathcal{E})$ is an assignment of one candidate to each entity mention $m_i$. $\mathcal{P}(*)$ is the power set operator.

The rank assignment and score calculation of the candidates of one entity is often not independent of the other entities' candidates. In this case, the ranking will be done by including the whole assignment via a global scoring function:
\begin{equation*}
 \mathit{score_{\mathit{global}}}: \mathcal{P}(\mathcal{E}) \to \mathbb{R} \text{ given by }A \mapsto \mathit{score_{\mathit{global}}}(A)
\end{equation*}
The objective function is then:
\begin{align*}
    A^* &= \argmax_A \left[\sum_{i=1}^k \mathit{score_{\mathit{local}}}(a_i, m_i)\right] \\ 
    &+ \mathit{score_{\mathit{global}}}(A) \ | \ a_i \in C_i
\end{align*}

Note, there also exists some ambiguity in the objective of linking itself. For example, there exists a Wikidata entity \texttt{2014 FIFA World Cup} and an entity \texttt{FIFA World Cup}. There is no unanimous solution on how to link the entity mention in the utterance \texttt{In 2014, Germany won the \underline{FIFA World Cup}}.

Sometimes EL is also called Entity Disambiguation, which we see more as part of EL, namely where entities are disambiguated via the candidate ranking.

There exist multiple special cases of EL.
\textit{Multilingual EL} tries to link entity mentions occurring in utterances of different languages to one shared KG, for example, English, Spanish or Chinese utterances to one language-agnostic KG. 
Formally, an entity mention $m$ in some utterance $u$ of some context language $l_c$ has to be linked to a language-agnostic KG which includes information in multiple languages $L_{KG}=\{l_1,...,l_k\}$ where $l_c$ can but has not to be an element of $L_{KG}$~\cite{Botha2020}.

\textit{Cross-lingual EL} tries to link entity mentions in utterances in different languages to a KG in one dedicated language, for example, Spanish and German utterances to an English KG~\cite{Rijhwani2019}. In that case, the multilingual EL problem gets constrained to $L_{KG}=\{l_{KG}\}$ where $l_c \neq l_{KG}$.

In \textit{zero-shot EL}, the entities during test time $\mathcal{E}_{\mathit{test}}$ are not available at training time $\mathcal{E}_{\mathit{train}}$. 
$$
\mathcal{E}_{\mathit{test}} \cap \mathcal{E}_{\mathit{train}} = \emptyset~\text{where}~\mathcal{E}_{\mathit{test}} \subset \mathcal{E},~\mathcal{E}_{\mathit{train}} \subset \mathcal{E}
$$
Thus, the entity linker must be able to handle unseen entities. 
The term was coined by Logeswaran et al.~\cite{Logeswaran2019}, but they limited the task to only have descriptions available while our definition does not include such a limitation. 

\textit{KB/KG-agnostic EL} approaches are able to support different KBs respectively KGs, often multiple in parallel. 
For example, a KG must be available in RDF format. We refer the interested reader to central works~\cite{DBLP:conf/ecai/UsbeckNRGCAB14, Moussallem2017, DBLP:conf/esws/ZwicklbauerSG16} or our Appendix.
\section{Wikidata}
\label{sec:wikidata}
Wikidata is a community-driven knowledge graph edited by humans and machines. 
The Wikidata community can enrich the content of Wikidata by, for example, adding/changing/removing entities, statements about them, and even the underlying ontology information.
As of July 2020, it contained around 87 million items of structured data about various domains. Seventy-three  million items can be interpreted as entities due to the existence of an \texttt{is instance} property. As a comparison, DBpedia contains around 5 million entities~\cite{Tanon2020}. Note that the \texttt{is instance} property includes a much broader scope of entities than the ones interpreted as entities for DBpedia.
In comparison to other similar KGs, the Wikidata dumps are updated most frequently~(\Cref{tab:kg_statistics}). But note that this only applies to the dumps, if one considers direct access via the Website or a SPARQL endpoint, both, Wikidata~\footnote{https://www.wikidata.org/}\footnote{https://query.wikidata.org} and DBpedia~\footnote{https://www.dbpedia.org/resources/live/}\footnote{https://www.dbpedia.org/resources/live/dbpedia-live-sync/} provide continuously updated knowledge.

\begin{table*}[htb!]
    \centering
    \begin{tabular}{l c  c c}
    \toprule
        \textbf{KG}& \textbf{\#Entities in million} & \textbf{\#Labels/Aliases in million} &\textbf{last updated} \\ \midrule
         Wikidata   & 78 & 442 & up to 4 times a month~\footnote{https://dumps.wikimedia.org/wikidatawiki/entities/} \\
         DBpedia    & 5 & 22 & monthly\\
         Yago4      & 67 & 371 & November 2019\\
         \bottomrule
    \end{tabular}
    \caption{KG statistics by~\cite{Tanon2020}.}
    \label{tab:kg_statistics}
\end{table*}

\subsection{Definition}
Wikidata is a collection of \emph{entities} where each such entity has a page on Wikidata. An entity can be either an \textit{item} or a \textit{property}. Note, an entity in the sense of Wikidata is generally not the same as an entity one links to via EL. For example, Wikidata entities are also properties that describe relations between different items. Linking to such relations is closer to Relation Extraction~\cite{Sorokin2017,Lin2017,Bastos2020}. Furthermore, many items are more abstract classes, which are usually not considered as entities linked to in EL.
Note that if not mentioned otherwise, if we speak about entities, entities in the context of EL are meant.
 
\begin{figure}[hb!]
    \includegraphics[width=\linewidth]{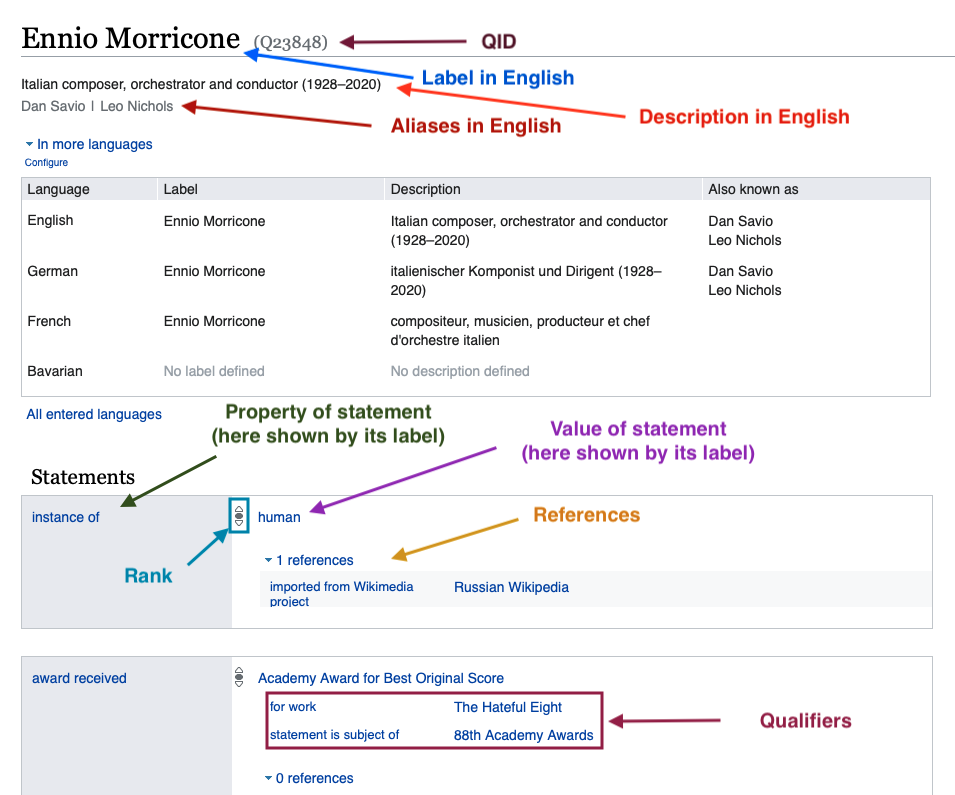}
    \caption{Example of an item in Wikidata.}
    \label{fig:item_example}
\end{figure}

\paragraph{Item.} Topics, classes, or objects are defined as items. An item is enriched with more information using statements about the item itself. In general, items consist of one label, one description, and aliases in different languages. An unique and language-agnostic identifier identifies items in the form \texttt{Q[0-9]+}. An example of an item can be found in Figure~\ref{fig:item_example}.

For example, the item with the identifier \texttt{Q23848} has the label \texttt{Ennio~Morricone}, two aliases, \texttt{Dan Savio} and \texttt{Leo Nichols}, and \texttt{Italian composer, orchestrator and conductor \\ (1928-2020)} as description at the point of writing. The corresponding Wikidata page can also be seen in Figure~\ref{fig:item_example}. 

\paragraph{Property.} A property specifies a relation between items or literals. Each property also has an identifier similar to an item, specified by \texttt{P[0-9]+}.  For instance, a property \texttt{P19} specifies the place of birth \texttt{Rome} for \texttt{Ennio~Morricone}. In NLP, the term \texttt{relation} is commonly used to refer to a connection between entities.  A property in the sense of Wikidata is a type of relation. To not break with the terminology used in the examined papers, when we talk about relations, we always mean Wikidata properties if not mentioned otherwise. 

\paragraph{Statement.} A statement introduces information by giving structure to the data in the graph. It is specified by a \emph{claim}, and \emph{references}, \emph{qualifiers} and \emph{ranks} related to the claim. Statements are assigned to items in Wikidata.
A claim is defined as a pair of property and some value. A value can be another item or some literal. Multiple values are possible for a property. Even an \texttt{unknown value} and a \texttt{no value} exists. 

\emph{References} point to sources making the claims inside the statements verifiable. In general, they consist of the source and date of retrieval of the claim. 

\emph{Qualifiers} define the value of a claim further by contextual information. For example, a qualifier could specify for how long one person was the spouse of another person. Qualifiers enable Wikidata to be hyper-relational (see \Cref{par:hyper}). Structures similar to qualifiers also exist in some other knowledge graphs, such as the inactive Freebase in the form of Compound Value Types~\cite{DBLP:conf/sigmod/BollackerEPST08}. 

\emph{Ranks} are used if multiple values are valid in a statement. If the population of a country is specified in a statement, it might also be useful to have the populations of past years available. The most up-to-date population information usually has then the highest rank and is thus usually the most desirable claim to use. 

Statements can be also seen in Figure~\ref{fig:item_example} at the bottom. For example, it is defined that \texttt{Ennio~Morricone} is an \texttt{instance of} the class \texttt{human}. This is also an example for the different types of items. While \texttt{Ennio~Morricone} is an entity in our sense, \texttt{human} is a class. 

\begin{figure*}[htb!]
    \begin{subfigure}{0.45\linewidth}
    \centering
    \includegraphics[width= \linewidth]{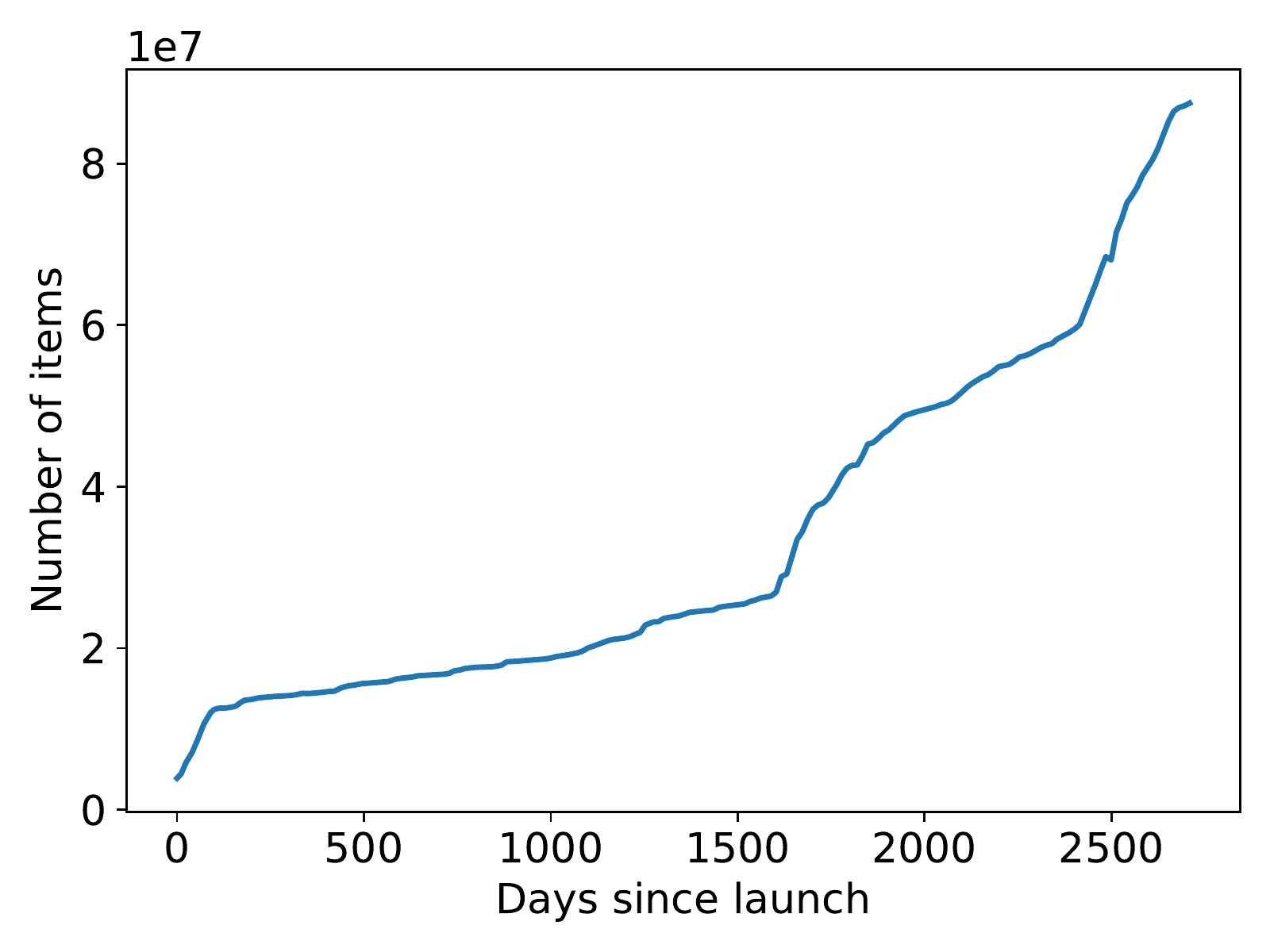}
    \caption{Number of items of Wikidata since launch~\cite{Manske2020}.}
    \label{fig:wikidata_items}
    \end{subfigure}\qquad
    \begin{subfigure}{0.45\linewidth}
      \centering
      \includegraphics[width=\linewidth]{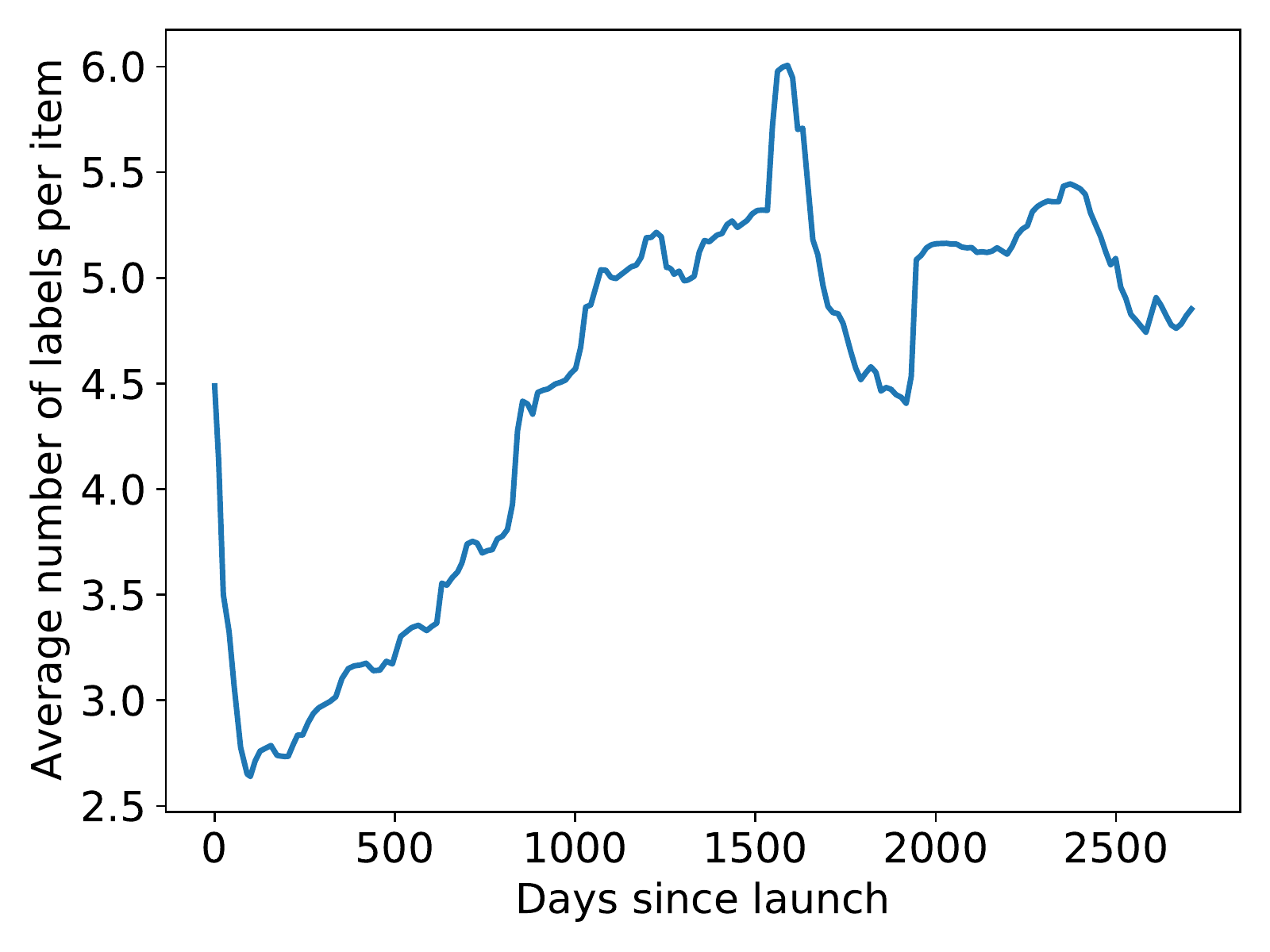}  
      \caption{Average number of labels (+ aliases) per item~\cite{Manske2020}.}
      \label{fig:sub_item_labels}
    \end{subfigure}
    
    \begin{subfigure}{0.45\linewidth}
      \centering
      \includegraphics[width=\linewidth]{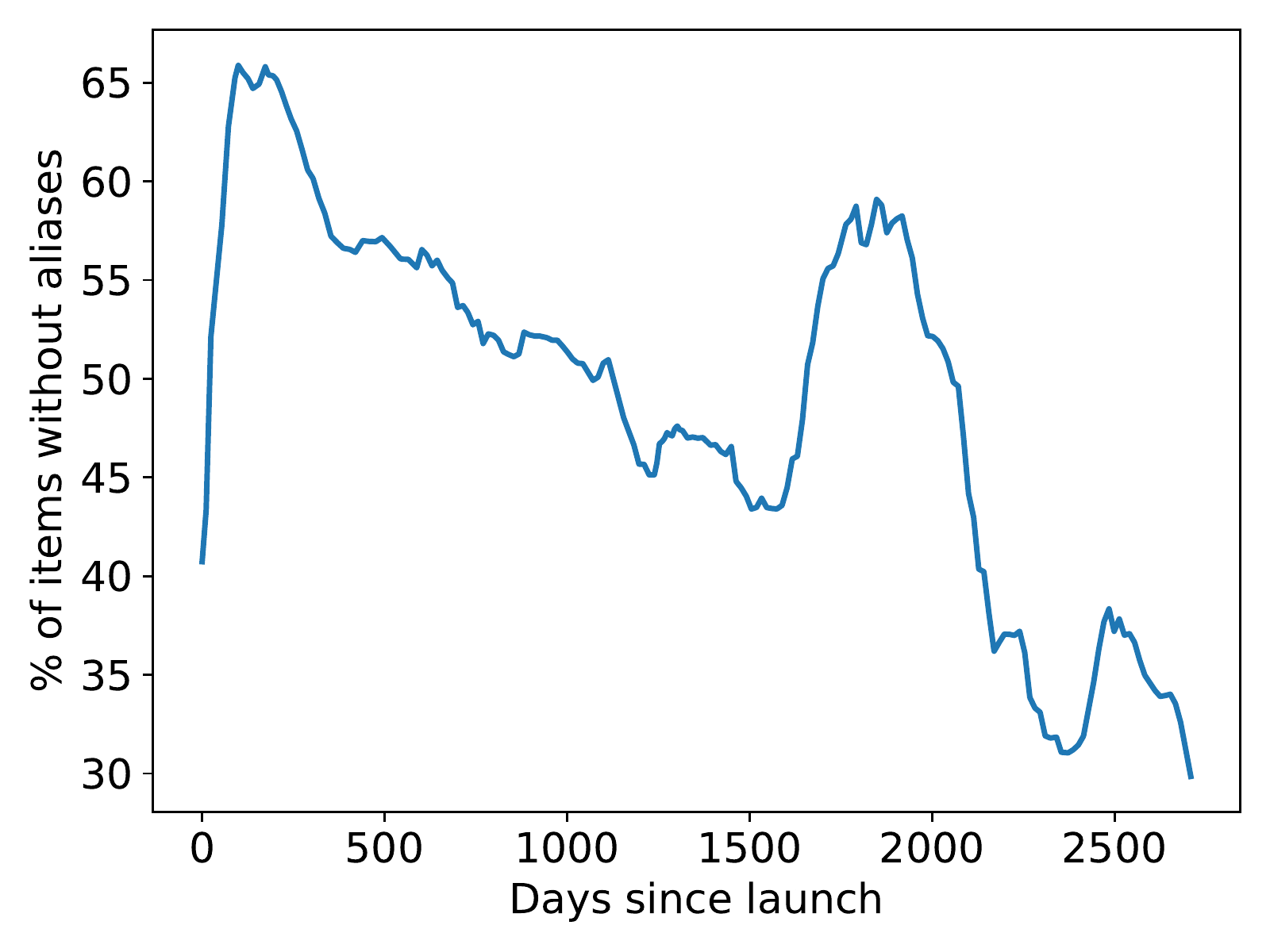}  
      \caption{Percentage of items without any aliases~\cite{Manske2020}.}
      \label{fig:item_labels_no_aliases}
    \end{subfigure}\qquad
    \begin{subfigure}{0.45\linewidth}
    \centering
    \includegraphics[width= \linewidth]{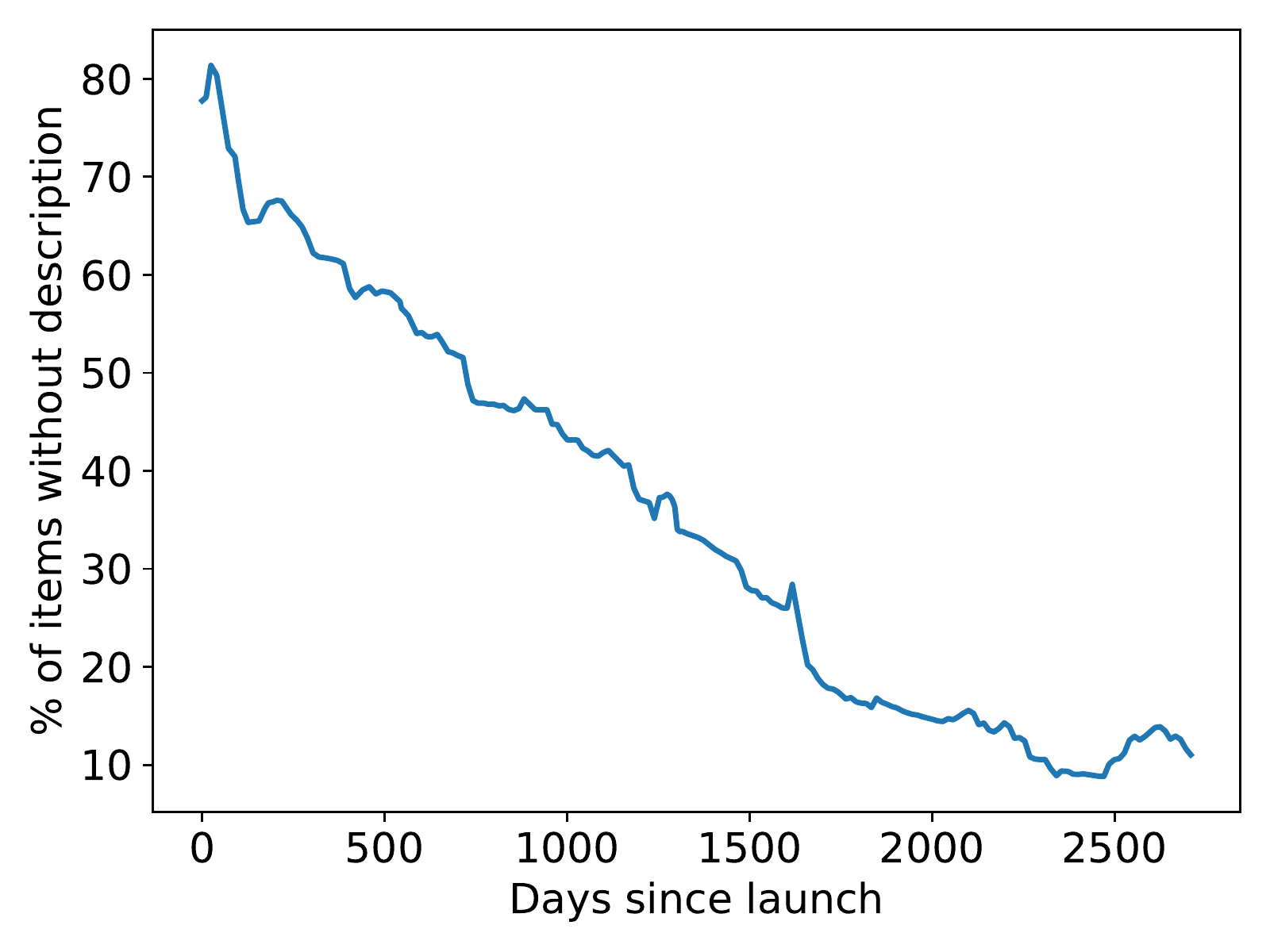}
    \caption{Percentage of items without a description~\cite{Manske2020}.}
    \label{fig:wikidata_descriptions}
     \end{subfigure}
    \caption{Statistics on Wikidata based on~\cite{Manske2020}.}
\end{figure*}

\paragraph{Other structural elements.}
The aforementioned elements are essential for Wikidata, but more do exist. For example, there are entities (in the sense of Wikidata) corresponding to Lexemes, Forms, Senses or Schemas. Lexemes, Forms and Senses are concerned with lexicographical information, hence words, phrases and sentences themselves. This is in contrast to Wikidata items and properties, which are directly concerned with things, concepts and ideas. Schemas formally subscribe to subsets of Wikidata entities. For example, any Wikidata item which has \texttt{actor} as its \texttt{occupation} is an \texttt{instance of} the class \texttt{human}. Both, lexicographical and schema information, are usually not directly of relevance for EL. Therefore, we refrain from introducing them in more detail.

For more information on Wikidata, see the paper by Denny Vrande{\v{c}}i{\'c} and Markus Kr{\"{o}}tzsch~\cite{DBLP:journals/cacm/VrandecicK14}.

\paragraph{Differences in structure to other knowledge graphs}
DBpedia extracts its information from Wikipedia and Wikidata. It maps the information to its own ontology.
DBpedia's statements consist of only single triples (\texttt{<subject, predicate, object>}) since it follows the RDF specification~\cite{DBLP:journals/semweb/LehmannIJJKMHMK15}. Additional information like qualifiers, references or ranks do not exist. But it can be modeled via additional triples. As it is no inherent feature of DBpedia, it is harder to use as there a no strict conventions. Entities in DBpedia have human-readable identifiers, and there exist entities per language~\cite{DBLP:journals/semweb/LehmannIJJKMHMK15} with partly differing information. Hence, for a single concept or thing, multiple DBpedia entities might exist. For example, the English entity of the city Munich\footnote{https://dbpedia.org/page/Munich} has 25 entities as \texttt{dbo:administrativeDistrict} assigned. The German entity\footnote{http://de.dbpedia.org/page/München} only a single one. It seems that this originates from a different interpretation of the predicate \texttt{dbo:administrativeDistrict}. 

Yago4 extracts all its knowledge from Wikidata but filters out information it deems inadequate. For example, if a property is used too seldom, it is removed. If a Wikidata entity does not have a class that exists in Schema.org\footnote{https://schema.org}, it is removed. The RDF specification format is used. Qualifier information is included indirectly via separate triples. Rank information and references of statements do not exist. The identifiers follow either a human-readable form if available via Wikipedia or Wikidata or use the Wikidata QID. However, in contrast to DBpedia, only one entity exists per thing or concept~\cite{Tanon2020}.

For a thorough comparison of Wikidata and other KGs (in respect to Linked Data Quality~\cite{zaveri2016quality}), please refer to the paper by Färber et al.~\cite{DBLP:journals/semweb/FarberBMR18}.
\subsection{Discussion}
\paragraph{Novelties.} A useful characteristic of Wikidata is that the community can openly edit it. Another novelty is that there can be a plurality of facts, as contradictory facts based on different sources are allowed. Similarly, time-sensitive data can also be included by qualifiers and ranks. The population of a country, for example, changes from year to year, which can be represented easily in Wikidata. Lastly, due to their language-agnostic identifiers, Wikidata is inherently multilingual. Language only starts playing a role in the labels and descriptions of an item.

\paragraph{Strengths.} Due to the inclusion of information by the community, recent events will likely be included. The knowledge graph is thus much more up-to-date than most other KGs. Freebase is unsupported for years now, and DBpedia updates its dumps only every month. Note, the novel DBpedia live 2.0\footnote{\url{https://forum.dbpedia.org/t/differences-in-results-on-dbpedia-live-and-http-dbpedia-org-sparql-endpoint/888/2}} is updated when changes to a Wikipedia page occur, but, as discussed, makes research harder to replicate. Thus, Wikidata is much more suitable and useful for industry applications such as smart assistants since it is the most complete open-accessible data source to date. In Figure~\ref{fig:wikidata_items}, one can see that number of items in Wikidata is increasing steadily. The existence of labels and additional aliases (see Figure~\ref{fig:sub_item_labels}) helps EL as a too-small number of possible surface forms often lead to a failure in the candidate generation. DBpedia does, for example, not include aliases, only a single exact label~\footnote{There exist some predicates (e.g., \texttt{foaf:name},  
\texttt{dbp:commonName} or \texttt{dbp:conventionalLongName}) that might point to aliases but they are often either not used or specify already stated aliases.}; to compensate, additional resources like Wikipedia are often used to extract a label dictionary of adequate size~\cite{Moussallem2017}. Even each property in Wikidata has a label~\cite{DBLP:journals/cacm/VrandecicK14}. Fully language model-based approaches are therefore more naturally usable~\cite{mulang2020encoding}.
Also, nearly all items have a description, see Figure~\ref{fig:wikidata_descriptions}. This short natural language phrase can be used for context similarity measures with the utterance. 
The inherent multilingual structure is intuitively useful for multilingual Entity Linking. 
Table~\ref{tab:statistics_languages} shows information about the use of different languages in Wikidata. As can be seen, item labels/aliases are available in up to 457 languages. But not all items have labels in all languages. On average, labels, aliases and descriptions are available in 29.04 different languages. However, the median is only 6 languages. Many entities will, therefore, certainly not have information in many languages. The most dominant language is English, but not all elements have label/alias/description information in English. For less dominant languages, this is even more severe. German labels exist, for example, only for 14 \%, and Samoan labels for 0.3 \%. 
Context information in the form of descriptions is also given in multiple languages. Still, many languages are again not covered for each entity (as can be seen by a median of only 4 descriptions per element). 
While the multilingual label and description information of items might be useful for language model-based variants, the same information for properties enables multilingual language models. Because, on average, 21.18 different languages are available per property for labels, one could train multilingual models on the concatenations of the labels of triples to include context information. But of course, there are again many properties with a lower number of languages, as the median is also only 6 languages. Cross-lingual EL is therefore certainly necessary to use language model-based EL in multiple languages.

\begin{table*}[htb!]
    \centering
    \begin{tabularx}{\linewidth}{p{10cm}XX}
         \toprule
         & \textbf{Items} & \textbf{Properties} \\
         \midrule
         \textbf{Number of languages} & 457 & 427\\
         \textbf{(average, median) of \# languages per element (labels +  descriptions)}&29.04, 6 &21.24, 13 \\
         \textbf{(average, median) of \# languages per element (labels)}&5.59 , 4 &21.18, 6 \\
         \textbf{(average, median) of \# languages per element (descriptions)}&26.10, 4&9.77, 6 \\
         \textbf{\% elements without English labels}&15.41\% & 0\% \\
         \textbf{\% elements without English descriptions}&26.23\%&1.08\% \\
         \bottomrule
    \end{tabularx}
    \caption{Statistics - Languages Wikidata (Extracted from dump~\cite{Foundation2020a}).}
    \label{tab:statistics_languages}
\end{table*}
By using the qualifiers of hyper-relational statements, more detailed information is available, useful not only for Entity Linking but also for other problems like Question Answering. The inclusion of hyper-relational statements is also more challenging. Novel graph embeddings have to be developed and utilized, which can represent the structure of a claim enriched with qualifiers~\cite{DBLP:conf/www/RossoYC20, Galkin2020}.  

Ranks are of use for EL in the following way. Imagine a person had multiple spouses throughout his/her life. In Wikidata, all those relationships are assigned to the person via statements of different ranks. If now an utterance is encountered containing information on the person and her/his spouse, one can utilize the Wikidata statements for comparison. Depending on the time point of the utterance, different statements apply. One could, for example, weigh the relevance of statements according to their rank. If now a KG (for example Yago4~\cite{Tanon2020}) includes only the most valid statement, the current spouse, utterances containing past spouses are harder to link. 

For references, up to now, no found approach did utilize them for EL. One use case might be to filter statements by reference if one knows the source's credibility, but this is more a measure to cope with the uncertainty of statements in Wikidata and not directly related to EL.

\begin{table*}[htb!]
    \centering
        \begin{tabular}{lccccc}
              \toprule
        \textbf{\# Labels/aliases}& \num{70124438} & \num{2041651} & \num{828471} & \num{89210} & \num{3329} \\
         \textbf{\# Items per label/alias}&$1$ & $2$  & $3-10$ & $11 - 100$ & $< 100$ \\ 
             \bottomrule
        \end{tabular}
    \caption{Number of English labels/aliases pointing to a certain number of items in Wikidata (Extracted from dump~\cite{Foundation2020a}).}
    \label{tab:mention_items}
\end{table*}

\paragraph{Weaknesses.} However, this community-driven approach also introduces challenges. For example, the list of labels of an item will not be exhaustive, as shown in Figures~\ref{fig:sub_item_labels} and ~\ref{fig:item_labels_no_aliases}. The graphs consider labels and aliases of all languages. While the median of labels and aliases is around 4 per element, not all are useful for Entity Linking. \texttt{Ennio Morricone} does not have an alias solely consisting of \texttt{Ennio} while he will certainly sometimes be referenced by that. Thus, one can not rely on the exact labels alone. But interestingly, Wikidata has properties for the fore- and surname alone, just not as a label or alias. A close examination of what information to use is essential. 

\begin{figure}[htb!]
    \centering
    \includegraphics[width=\linewidth]{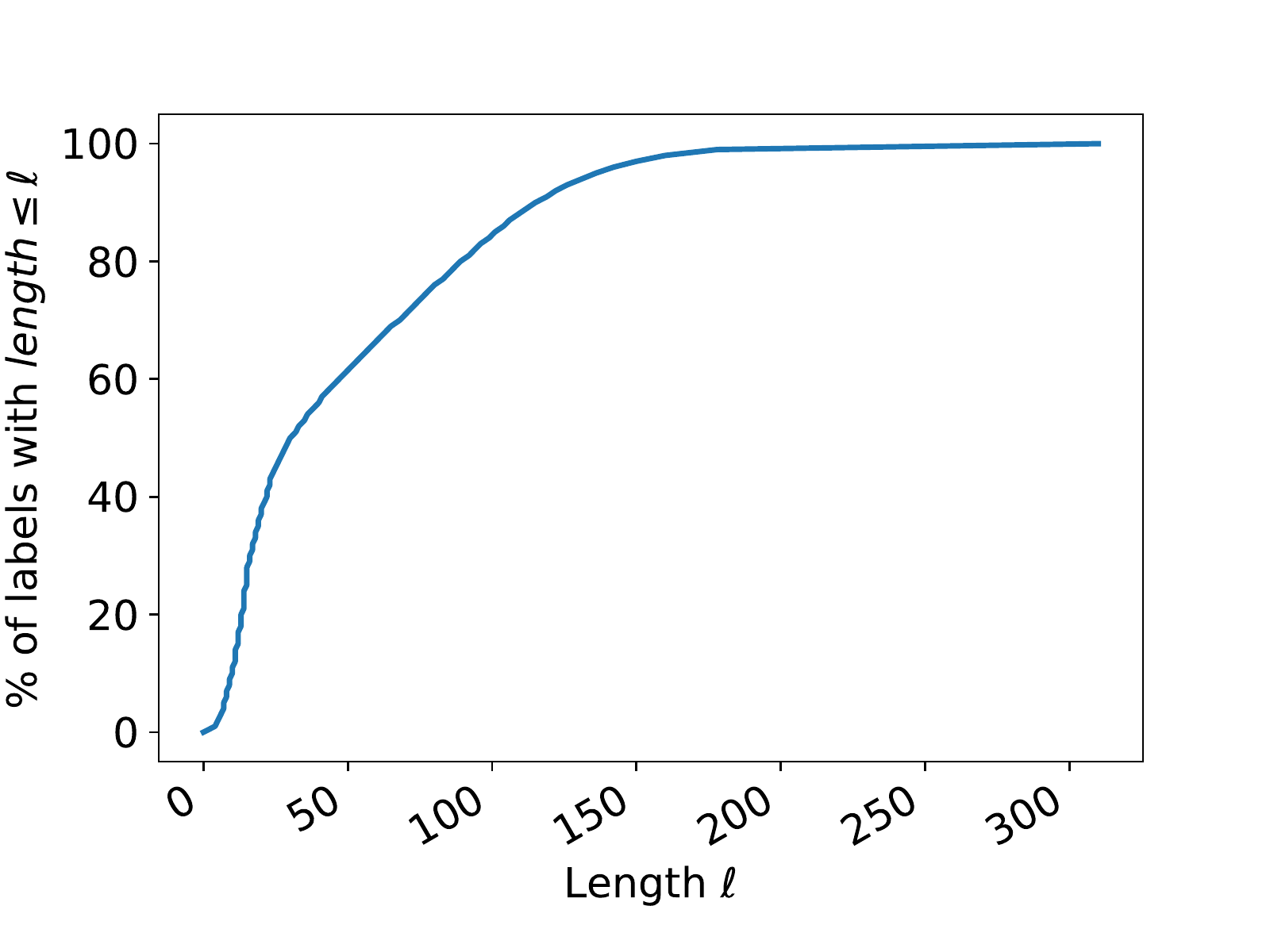}
    \caption{Percentiles of English label lengths (Extracted from dump~\cite{Foundation2020a}).}
    \label{fig:wikidata_label_lengths}
\end{figure}

This is also a problem in other KGs.  Also, Wikidata often has items with very long, noisy, error-prone labels, which can be a challenge to link to~\cite{mulang2020encoding}. Nearly 20 percent of labels have a length larger than 100 letters, see Figure~\ref{fig:wikidata_label_lengths}. Due to the community-driven approach, false statements also occur due to errors or vandalism~\cite{Heindorf2016}.

Another problem is that entities lack of facts (here defined as statements not being labels, descriptions, or aliases). According to Tanon et al.~\cite{Tanon2020}, in March 2020, DBpedia had, on average, $26$ facts per entity while Wikidata had only $12.5$. This is still more than YAGO4 with $5.1$. 
To tackle such long-tail entities, different approaches are necessary.
The lack of descriptions can also be a problem. Currently, around 10\% of all items do not have a description, as shown in Figure~\ref{fig:wikidata_descriptions}. Luckily, the situation is increasingly improving. 

A general problem of Entity Linking is that a label or alias can reference multiple entities, see Table~\ref{tab:mention_items}. While around 70 million mentions point each to a unique item, 2.9 million do not. Not all of those are entities by our definition but, e.g., also classes or topics. In addition, longer labels or aliases often correspond to non-entity items. Thus, the percentage of entities with overlapping labels or aliases is certainly larger than for all items. To use Wikidata as a Knowledge Graph, one needs to be cautious of the items one will include as entities. For example, there exist \texttt{Wikimedia disambiguation page} items that often have the same label as an entity in the classic sense. Both \texttt{Q76} and \texttt{Q61909968} have \texttt{Barack Obama} as the label. Including those will make disambiguation more difficult.
Also, the possibility of contradictory facts will make EL over Wikidata harder. 

In Wikification, also known as EL on Wikipedia, large text documents for each entity exist in the knowledge graph, enabling text-heavy methods~\cite{Wu2019}. Such large textual contexts (besides the descriptions and the labels of triples itself) do not exist in Wikidata, requiring other methods or the inclusion of Wikipedia. However, as Wikidata is closely related to Wikipedia, an inclusion is easily doable. Every Wikipedia article is connected to a Wikidata item. The Wikipedia article belonging to a Wikidata item can be, for example, extracted via a SPARQL~\footnote{https://www.w3.org/TR/rdf-sparql-query/} query to the Wikidata Query Service~\footnote{https://query.wikidata.org} using the \texttt{http://schema.org/about} predicate. The Wikidata item of a Wikipedia article can be simply found on the article page itself or by using the Wikipedia API~\footnote{https://en.wikipedia.org/w/api.php}. 

One can conclude that the characteristics of Wikidata, like being up to date, multilingual and hyper-relational, introduce new possibilities. At the same time, the existence of long-tail entities, noise or contradictory facts poses a challenge.

\section{Datasets}
\label{sec:datasets}
\subsection{Overview}
This section is concerned with analyzing the different datasets which are used for Wikidata EL. A comparison can be found in Table~\ref{tab:datasets}. 
\begin{table*}[ptb!]
    \small
    \centering
    \rotatebox{-90}{
    \begin{threeparttable}
    \begin{tabularx}{0.95\textheight}{p{4cm}p{3cm}XXp{2.8cm}P{1.5cm}p{3cm}}
         \toprule
         \textbf{Dataset} & \textbf{Domain} & \textbf{Year} & \textbf{Annotation process} & \textbf{Purpose} &  \textbf{Spans given} & \textbf{Identifiers} \\ \midrule
         T-REx~\cite{elsahar2019t} & Wikipedia abstracts  & 2015& automatic & Knowledge Base Population (KBP), Relation Extraction (RE), Natural Language Generation (NLG) &\cmark& Wikidata \\ 
         NYT2018~\cite{DBLP:journals/pvldb/LinLXLC20, DBLP:conf/icde/LinC19} & News & 2018& manually & EL &\cmark & Wikidata, DBpedia \\
         ISTEX-1000~\cite{DBLP:journals/corr/abs-1904-09131} & Research articles& 2019 & manually&EL & \cmark& Wikidata \\
         LC-QuAD 2.0~\cite{dubey2019lc} & General  complex questions (Wikidata) & 2019& semi-automatic &Question Answering (QA) &\xmark& DBpedia, Wikidata \\
         Knowledge Net~\cite{mesquita2019knowledgenet} & Wikipedia abstracts, biographical texts & 2019& manually & KBP  &\cmark& Wikidata
         \\
         KORE50DYWC~\cite{noullet2020kore} & News & 2019& manually & EL & \cmark& Wikidata, DBpedia, YAGO, Crunchbase \\
         Kensho Derived Wikimedia Dataset~\cite{KenshoRD} & Wikipedia & 2020& automatic & Natural Language Processing (NLP) &\cmark & Wikidata, Wikipedia  \\
         CLEF HIPE 2020~\cite{Ehrmann2020} & Historical newspapers & 2020& manually & ER, EL & \cmark &Wikidata  \\
         Mewsli-9~\cite{Botha2020} & News in multiple languages & 2020& automatic & Multilingual EL & \cmark & Wikidata  \\
         TweekiData~\cite{tweeki:wnut20} & Tweets&2020& automatic&EL&\cmark & Wikidata \\
         TweekiGold~\cite{tweeki:wnut20}&Tweets&2020& manually&EL&\cmark& Wikidata\\
         \bottomrule
    \end{tabularx}
    \begin{tablenotes}
        \setlength{\columnsep}{0.8cm}
        \setlength{\multicolsep}{0cm}
        \begin{multicols}{2}
            \item[1] \label{datasets:tnote1} data from 2010
            \item[2] \label{datasets:tnote2} Original dataset on Wikipedia
        \end{multicols}
    \end{tablenotes}
    \end{threeparttable}
    }
    \caption{Comparison of used datasets.}
    \label{tab:datasets}
\end{table*}
The majority of datasets on which existing Entity linkers were evaluated,  were originally constructed for KGs different from Wikidata. Such a mapping can be problematic as some entities labeled for other KGs could be missing in Wikidata. Or some NIL entities that do not exist in other KGs could exist in Wikidata.
Eleven datasets~\cite{DBLP:journals/corr/abs-1904-09131,DBLP:journals/pvldb/LinLXLC20,dubey2019lc, elsahar2019t, noullet2020kore, mesquita2019knowledgenet, KenshoRD, Ehrmann2020, Botha2020, tweeki:wnut20} were found for which Wikidata identifiers were available from the start. 
In the following the datasets are separated by their domain. A list of all examined datasets - including links where available - can be found in the Appendix in \Cref{tab:dataset_links}.

\subsubsection{Encyclopedic datasets}
LC-QuAD 2.0~\cite{dubey2019lc} is a semi-automatically created dataset  for  Questions Answering providing complex natural language questions. For each question, Wikidata and DBpedia identifiers are provided. The questions are generated from subgraphs of the Wikidata KG and then manually checked. The dataset does not provide annotated mentions.

T-REx~\cite{elsahar2019t} was constructed automatically over Wiki\-pedia abstracts. Its main purpose is Knowledge Base Population (KBP). According to Mulang et al.~\cite{mulang2020encoding}, this dataset describes the challenges of Wikidata, at least in the form of long, noisy labels, the best. 

The Kensho Derived Wikimedia Dataset~\cite{KenshoRD} is an automatically created condensed subset of Wikimedia data. It consists of three levels: Wikipedia text, annotations with Wikipedia pages and links to Wikidata items. Thus, mentions in Wikipedia articles are annotated with Wikidata items. However, as some Wikidata items do not have a corresponding Wikipedia page, the annotation is not exhaustive. It was constructed for NLP in general. 

\subsubsection{Research-focused datasets}
ISTEX-1000~\cite{DBLP:journals/corr/abs-1904-09131} is a research-focused dataset containing 1000 author affiliation strings. It was manually annotated to evaluate the OpenTapioca~\cite{DBLP:journals/corr/abs-1904-09131} entity linker. 

\subsubsection{Biographical datasets}
KnowledgeNet~\cite{mesquita2019knowledgenet} is a Knowledge Base Population dataset with 9073 manually annotated sentences. The text was extracted from biographical documents from the web or Wikipedia articles.

\subsubsection{News datasets}
NYT2018~\cite{DBLP:journals/pvldb/LinLXLC20, DBLP:conf/icde/LinC19} consists of 30 news documents that were manually annotated on Wikidata and DBpedia. It was constructed for KBPearl~\cite{DBLP:journals/pvldb/LinLXLC20}, so its main focus is also KBP which is a downstream task of EL.

One dataset, KORE50DYWC~\cite{noullet2020kore}, was found, which was not used by any of the approach papers. 
It is an annotated EL dataset based on the KORE50 dataset, a manually annotated subset of the AIDA-CoNLL corpus. The original KORE50 dataset focused on highly ambiguous sentences. All sentences were reannotated with DBpedia, Yago, Wikidata and Crunchbase entities.

CLEF HIPE 2020~\cite{Ehrmann2020} is a dataset based on historical newspapers in English, French and German. Only the English dataset will be analyzed in the following. This dataset is of great difficulty due to many errors in the text, which originate from the OCR method used to parse the scanned newspapers. For the English language, only a development and test set exist. In the other two languages, a training set is also available. It was manually annotated.

Mewsli-9~\cite{Botha2020} is a multilingual dataset automatically constructed from WikiNews. It includes nine different languages. A high percentage of entity mentions in the dataset do not have corresponding English Wikipedia pages, and thus, cross-lingual linking is necessary. Again, only the English part is included during analysis.

\subsubsection{Twitter datasets}
TweekiData and TweekiGold~\cite{tweeki:wnut20} are an automatically annotated corpus and a manually annotated dataset for EL over tweets. TweekiData was created by using other existing tweet-based datasets and linking them to Wikidata data via the Tweeki EL. TweekiGold was created by an expert, manually annotating tweets from another dataset with Wikidata identifiers and Wikipedia page-titles. 

\subsection{Analysis} 
\begin{table*}[ptb!]
    \centering
    \rotatebox{-90}{
    \begin{threeparttable}
        \begin{tabularx}{0.95\textheight}{p{5cm}p{2cm}p{2cm}XXXX}
            \toprule
             \textbf{Dataset}& \# \textbf{documents} & \# \textbf{mentions} & \textbf{NIL entities} & \textbf{Wikidata entities} & \textbf{Unique Wikidata entities} & \textbf{Mentions per document} \\ \midrule
             T-REx~\cite{elsahar2019t} & \num{4650000} & \num{51297484}&0\% &100\% & 9.1\% & 11.03\\
             NYT2018~\cite{DBLP:journals/pvldb/LinLXLC20, DBLP:conf/icde/LinC19}~\tnotex{tn:tabced1} & \num{30}
             &- &-& - & - & -\\
             ISTEX-1000~\cite{DBLP:journals/corr/abs-1904-09131} (train) & \num{750}& 2073& 0\%&100\%& 53.7\%& 2.76\\
             ISTEX-1000~\cite{DBLP:journals/corr/abs-1904-09131} (test) & 250 & 670 & 0\%&100\%& 65.8\%&2.68\\
             LC-QuAD 2.0~\cite{dubey2019lc} & \num{6046} & \num{44529} & 0\%& 100\%& 51.2\% & 1.47\\
             Knowledge Net~\cite{mesquita2019knowledgenet} (train) & \num{3977} & \num{13039}& 0\%&100\%& 30\% &3.28\\
             Knowledge Net~\cite{mesquita2019knowledgenet} (test)~\tnotex{tn:tabced2} & \num{1014} & -& -&-& - &-\\
             KORE50DYWC~\cite{noullet2020kore} & \num{50} & \num{307}& 0\%&100\%  & 72.0\%  & 6.14 \\
             Kensho Derived Wikimedia Dataset~\cite{KenshoRD} & \num{14255258} &\num{121835453}&0\% &100\%&3.7\%& 8.55\\
             CLEF HIPE 2020 (en, dev)~\cite{Ehrmann2020} & 80& 470& 46.4\%&53.6\%& 31.9\%& 5.88\\
             CLEF HIPE 2020 (en, test)~\cite{Ehrmann2020} & 46&134& 33.6\%&66.4\%& 42.5\% & 2.91\\
             Mewsli-9 (en)~\cite{Botha2020} & \num{12679}  
             &\num{80242} &0\% & 100\% & 48.2\% &6.33\\
             TweekiData~\cite{tweeki:wnut20}&\num{5000000}&\num{5038870}& 61.2\%& 38.8\%&5.4\% &1.01\\
             TweekiGold~\cite{tweeki:wnut20}&500&\num{958}&11.1\%&88.9\%&66.6\%&1.92\\
             \bottomrule
        \end{tabularx}
        \begin{tablenotes}
            \item[1] \label{tn:tabced1} Information gathered from accompanying paper as dataset was not available 
            \item[2] \label{tn:tabced2} Available dataset did not contain mention/entity information
        \end{tablenotes}
        \end{threeparttable}
    }
    \caption{Comparison of the datasets with focus on the number of documents and Wikidata entities.}
    \label{tab:comp_entities_datasets}
\end{table*}
Table~\ref{tab:comp_entities_datasets} shows the number of documents, the number of mentions, NIL entities and unique entities, and the mentioned ratio. What classifies as a document in a dataset depends on the dataset itself. For example, for T-REx, a document is a whole paragraph of a Wikipedia article, while for LC-QuAD 2.0, a document is just a single question. Due to this, the average number of entities in a document also varies, e.g., LC-QuAD 2.0 with $1.47$ entities per document and T-REx with $11.03$. If a dataset was not available, information from the original paper was included. If dataset splits were available, the statistics are also shown separately. The majority of datasets do not contain NIL entities. For the Tweeki datasets, it is not mentioned which Wikidata dump was used to annotate. For a dataset that contains NIL entities, this is problematic. On the other hand, the dump is specified for the CLEF HIPE 2020 dataset, making it possible to work on the Wikidata version with the correct entities missing.

\begin{highlightbox}{\hyperlink{rq1}{Research Question 1}}{Which Wikidata EL datasets exist, how widely used are they and how are they constructed?}
The preceding paragraphs answer the following two aspects of the first research question. First, we provided descriptions and an overview of all datasets created for Wikidata, including statistics on their structure. This answers which datasets exist. Furthermore, for each dataset it is stated how they were constructed, whether automatically, semi-automatically or manually. Thus information on the quality and construction process of the datasets is given.
To answer the last part of the question, how widely are the datasets in use, Table~\ref{tab:dataset_usage} shows how many times each Wikidata dataset was used in Wikidata EL approaches during training or evaluation. As one can see, there exists no single dataset used in all research of EL. This is understandable as different datasets focus on different document types and domains as shown in Table~\ref{tab:datasets}, what again results in different approaches. 
\end{highlightbox}

\begin{table*}[htb!]
    \centering
    \begin{threeparttable}
        \begin{tabular}{lc}
        \toprule
        \textbf{Dataset} & \textbf{Number of usages in Wikidata EL approach papers} \\
        \midrule
        T-REx~\cite{elsahar2019t} & 2~\cite{mulang2020encoding, DBLP:journals/pvldb/LinLXLC20} \\
        NYT2018~\cite{DBLP:journals/pvldb/LinLXLC20, DBLP:conf/icde/LinC19} & 1~\cite{DBLP:journals/pvldb/LinLXLC20}
        \\
        ISTEX-1000~\cite{DBLP:journals/corr/abs-1904-09131} & 2~\cite{DBLP:journals/corr/abs-2008-05190, DBLP:journals/corr/abs-1904-09131} \\
        LC-QuAD 2.0~\cite{dubey2019lc} & 2~\cite{DBLP:journals/corr/abs-1912-11270, Banerjee2020} \\
        Knowledge Net~\cite{mesquita2019knowledgenet} & 1~\cite{DBLP:journals/pvldb/LinLXLC20}
        \\
        KORE50DYWC~\cite{noullet2020kore} & 0 \\
        Kensho Derived Wikimedia Dataset~\cite{KenshoRD} & 1~\cite{perkins2020separating} \\
        CLEF HIPE 2020~\cite{Ehrmann2020} & 3~\cite{borosrobust, Labusch2020, Provatorova2020}  \\
        Mewsli-9~\cite{Botha2020} & 1~\cite{Botha2020}  \\
        TweekiData~\cite{tweeki:wnut20} & 1~\cite{tweeki:wnut20} \\
        TweekiGold~\cite{tweeki:wnut20}& 1~\cite{tweeki:wnut20}\\
        \bottomrule
    \end{tabular}
    \caption{Usage of datasets for training or evaluation.}
    \label{tab:dataset_usage}
    \end{threeparttable}
\end{table*}

\begin{table*}[htb!]
    \centering
    \begin{tabularx}{\textwidth}{p{5cm}XXXX}
     \toprule
     \textbf{Dataset} & \textbf{Average number of matches} & \textbf{No match} & \textbf{Exact match} & \textbf{More than one match} \\ \midrule
     T-REx & 4.79 & 31.36\% & 32.98\% & 35.65\% \\
     ISTEX-1000 (train) & 23.23 & 8.06\%& 26.34\% & 65.61\%\\
     ISTEX-1000 (test) & 25.85 & 10.30\% & 23.88\% & 65.82\%\\
     Knowledge Net (train) & 21.90 & 10.41\% & 22.29\% & 67.3\% \\
     KORE50DYWC & 28.31 & 3.93\% & 7.49\% & 88.60\%  \\
     Kensho Derived Wikimedia Dataset & 8.16 & 35.18\% & 30.94\% & 33.88\% \\
     CLEF HIPE 2020 (en, dev) & 24.02 & 35.71\% & 11.51\% & 52.78\% \\
     CLEF HIPE 2020 (en, test) & 17.78 & 43.82\% & 6.74\% & 49.44\% \\
     Mewsli-9 (en) &11.09&16.80\%&34.90\%&47.30\% \\
     TweekiData &19.61&19.98\%&12.01\%&68.01\% \\
     TweekiGold &16.02&7.41\%&20.25\%&72.34\% \\
     \bottomrule
    \end{tabularx}
    \caption{Ambiguity of mentions (existence of a match does not correspond to a correct match), NYT2018 dataset was not available and LC-QuAD 2.0 is not annotated.}
    \label{tab:ambiguity_mentions}
\end{table*}
\begin{table}[htb!]
    \centering
    \begin{tabularx}{\linewidth}{p{4cm}cc}
     \toprule
     \textbf{Dataset} & \textbf{Acc.} & \textbf{Acc. filtered} \\ \midrule
     ISTEX-1000 (train) & 0.744 &0.716\\
     ISTEX-1000 (test) & 0.716&0.678\\
     Knowledge Net (train) &  0.371&0.285 \\
     KORE50DYWC & 0.225& 0.187\\
     CLEF HIPE 2020 (en, dev) & 0.333&0.287 \\
     CLEF HIPE 2020 (en, test) & 0.258& 0.241\\
     TweekiGold & 0.565 & 0.520\\
     Mewsli-9 (en) & 0.602&0.490\\
     \bottomrule
    \end{tabularx}
    \caption{EL accuracy - Kensho Derived Wikimedia Dataset, T-REx and TweekiData are not included due to size, \textbf{Acc. filtered} has all exact matches removed, NYT2018 dataset was not available and LC-QuAD 2.0 is not annotated.}
    \label{tab:es}
\end{table}
The difficulty of the different datasets was measured by the accuracy of a simple EL method (Table~\ref{tab:es}) and the ambiguity of mentions (Table~\ref{tab:ambiguity_mentions}). 
The simple EL method searches for entity candidates via an ElasticSearch index, including all English labels and aliases. It then disambiguates by taking the one with the largest tf-idf-based BM25 similarity measure score and the lowest Q-identifier number resembling the popularity. Nothing was done to handle inflections.\footnote{All source code, plots and results can be found on \url{https://github.com/semantic-systems/ELEnglishWD}}
Only accessible datasets were included. As one can see, the accuracy is positively correlated with the number of exact matches. The more ambiguous the underlying entity mentions are, the more inaccurate a simple similarity measure between label and mention becomes. In this case, more context information is necessary. The simple Entity Linker was only applied to datasets that were feasible to disambiguate in that way. T-REx and the Kensho Derived Wikimedia Dataset were too large in terms of the number of documents to run the linker on commodity hardware. 
According to the EL performance, ISTEX-1000 is the easiest dataset. Many of the ambiguous mentions reference the most popular one, while also many exact unique matches exist.
T-REx, the Kensho Derived Wikimedia Dataset and the Mewsli-9 training dataset have the largest percentage of exact matches for labels.
While TweekiGold is quite ambiguous, deciding on the most prominent entity appears to produce good EL results.
The most ambiguous dataset is KORE50DYWC. Additionally, just choosing the most popular entity of the exact matches results in worse performs than for example on TweekiGold which is also very ambiguous.  This is due to the fact that the original KORE50 dataset focuses on difficult ambiguous entities which are not necessarily popular. The CLEF HIPE 2020 dataset also has a low EL accuracy but not due to ambiguity but many mentions with no exact match.  The reason for that is the noise created by OCR. 

The second column of Table~\ref{tab:es} specifies the accuracy with all unique exact matches removed. This is based on the intuition that exact matches without any competitors are usually correct. 

As seen in the \Cref{tab:datasets,tab:ambiguity_mentions,tab:comp_entities_datasets,tab:es}, there exists a very diverse set of datasets for EL on Wikidata, differing in the domain, document type, ambiguity and difficulty. 
\begin{highlightbox}{\hyperlink{rq2}{Research Question 2}}{Do the characteristics of Wikidata matter for the design of EL datasets and if so, how?}
Except the Mewsli-9~\cite{Botha2020} and CLEF HIPE 2020~\cite{Ehrmann2020} datasets, none of the others take any specific characteristics of Wikidata into account. 
The two exceptions focus on multilinguality and rely therefore directly on the language-agnostic nature of Wikidata.
The CLEF HIPE 2020 dataset is designed for Wikidata and has documents for  English, French and German, but each language has a different corpus of documents. The same is the case for the Mewsli-9 dataset, while here, documents in nine languages are available.
In the future, a dataset similar to VoxEL~\cite{rosales2018voxel}, which is defined for Wikipedia, would be helpful. Here, each utterance is translated into multiple languages, which eases the comparison of the multilingual EL performance. Having the same corpus of documents in different languages would allow a better comparison of a method's performance in various languages. Of course, such translations will never be perfectly comparable. 
\end{highlightbox}

Besides that, we identified one additional characteristic which might be of relevance to Wikidata EL datasets.
It is the large rate of change of Wikidata.
Due to that, it would be advisable that the datasets specify the Wikidata dumps they were created on, similar to Petroni et al.~\cite{petroni2020kilt}. Many of the existing datasets do that, yet not all. In current dumps, entities, which were available while the dataset was created, could have been removed. It is even more probable that NIL entities could now have a corresponding entity in an updated Wikidata dump version. If the EL approach now would detect it as a NIL entity, it is evaluated as correct, but in reality, this is false and vice versa. 
Of course, this is not a problem unique to Wikidata. Anytime, the dump is not given for an EL dataset, similar uncertainties will occur. But due to the fast growth of Wikidata (see Figure~\ref{fig:wikidata_items}), this problem is more pronounced.

Concerning \textit{emerging entities}, another variant of an EL dataset could be useful too. Two Wikidata dumps from different time points could be used to label the utterances. Such a dataset would be valuable in the context of an EL approach supporting emerging entities (e.g., the approach by Hoffart et al.~\cite{DBLP:conf/www/HoffartAW14}). With the true entities available, one could measure the quality of the created emerging entities. That is, multiple mentions assigned to the same emerging entity should also point to a single entity in the more recent KG. 
Also, constraining that the method needs to perform well on both KG dumps would force EL approaches to be less reliant on a fixed graph structure.

\section{Approaches}
\label{sec:approaches}
Currently, the number of methods intended to work explicitly on Wikidata is still relatively small, while the amount of the ones utilizing the characteristics of Wikidata is even smaller. 

There exist several KG-agnostic EL approaches~\cite{Moussallem2017,DBLP:conf/ecai/UsbeckNRGCAB14,DBLP:conf/esws/ZwicklbauerSG16}. However, they were omitted as their focus is being independent of the KG. While they are able to use Wikidata characteristics like labels or descriptions, there is no explicit usage of those. They are available in most other KGs. None of the found KG-agnostic EL papers even mentioned Wikidata. Though we recognize that KG-agnostic approaches are very useful in the case that a KG becomes obsolete and has to be replaced or a non-public KG needs to be used, such approaches are not included in this section. However, Table~\ref{tab:comparison_KG_agnostic_approaches_wikidata} in the Appendix provides an overview of the used Wikidata characteristics of the three approaches. 

DeepType~\cite{DBLP:conf/aaai/RaimanR18} is an entity linking approach relying on the fine-grained type system of Wikidata and the categories of Wikipedia. As type information is not evolving as fast as novel entities appear, it is relatively robust against a changing knowledge base. While it uses Wikidata, it is not specified in the paper whether it links to Wikipedia or Wikidata. Even the examination of the available code did not result in an answer as it seems that the entity linking component is missing. While DeepType showed that the inclusion of Wikidata type information is very beneficial in entity linking, we did not include it in this survey due to the aforementioned reasons.
As Wikidata contains many more types ($\approx$\num{2400000}) than other KGs, e.g., DBpedia ($\approx$\num{484000})~\cite{Tanon2020}~\footnote{if all rdf:type objects are considered, else $\approx$ 768 (gathered via https://dbpedia.org/sparql/) if only considering types of the DBpedia ontology}), it seems to be more suitable for this fine-grained type classification. Yet, not only the number of types plays a role but also how many types are assigned per entity. In this regard, Wikipedia provides much more type information per entity than Wikidata~\cite{Weikum2020}. That is probably the reason why both Wikipedia categories and Wikidata types are used together. As Wikidata is growing every minute, it may also be challenging to keep the type system up to date. 

Tools without accompanying publications are not considered due to the lack of information about the approach and its performance. Hence, for instance, the Entity Linker in the 
DeepPavlov~\cite{burtsev2018deeppavlov} 
framework is not included, although it targets Wikidata and appears to use label and description information successfully to link entities.

While the approach by Zhou et al.~\cite{DBLP:journals/tacl/ZhouRWCN20} does utilize Wikidata aliases in the candidate generation process, the target KB is Wikipedia and was therefore excluded.

The vast majority of methods is using machine learning to solve the EL task~\cite{DBLP:journals/corr/abs-1810-09164,mulang2020encoding,DBLP:conf/lrec/KlangN20,DBLP:conf/starsem/SorokinG18, Banerjee2020, huangentity, perkins2020separating, DBLP:journals/corr/abs-2008-05190, borosrobust, Provatorova2020, Labusch2020, Botha2020, DBLP:journals/corr/abs-1904-09131}. Some of those approaches solve the ER and EL jointly as an end-to-end task.  Besides that, there exist two rule-based approaches
~\cite{DBLP:journals/corr/abs-1912-11270, tweeki:wnut20} and two based on graph optimization~\cite{DBLP:conf/lrec/KlangN20, DBLP:journals/pvldb/LinLXLC20}.

The approaches mentioned above solve the EL problem as specified in Section~\ref{sec:problem}. That is, other EL methods with a different problem definition also exist. For example, Almeida et al.~\cite{almeida2016streets} try to link street names to entities in Wikidata by using additional location information and limiting the entities only to locations. As it uses additional information about the true entity via the location, it is less comparable to the other approaches and, thus, was excluded from this survey.  Thawani et al.~\cite{DBLP:conf/semweb/ThawaniHHZDSQSP19} link entities only over columns of tables. The approach is not comparable since it does not use natural language utterances. 
The approach by Klie et al.~\cite{DBLP:conf/acl/KlieCG20} is concerned with Human-In-The-Loop EL. While its target KB is Wikidata, the focus on the inclusion of a human in EL process makes it incomparable to the other approaches.
EL methods exclusively working on languages other than English~\cite{ElVaigh2020,Ellgren2020, Klang2014, Veen2016, Ehrmann2020a} were not considered but also did not use any novel characteristics of Wikidata.
In connection to the CLEF HIPE 2020 challenge~\cite{Ehrmann2020a}, multiple Entity Linkers working on Wikidata were built. While short descriptions of the approaches are available in the challenge-accompanying paper, only approaches described in an own published paper were included in this survey. The approach by Kristanti and Romary~\cite{Kristanti2020} was not included as it used pre-existing tools for EL over Wikidata, for which no sufficient documentation was available. 

Due to the limited number of methods, we also evaluated methods that are not solely using Wikidata but also additional information from a separate KG or Wikipedia. This is mentioned accordingly. Approaches linking to knowledge graphs different from Wikidata, but for which a mapping between the knowledge graphs and Wikidata exists, are also not included. Such methods would not use the Wikidata characteristics at all, and their performance depends on the quality of the other KG and the mapping.

In the following, the different approaches are described and examined according to the used characteristics of Wikidata. An overview can be found in \Cref{tab:comparison_approaches_wikidata}.
We split the approaches into two categories, the ones doing only EL and the ones doing ER and EL.
Furthermore, to provide a better overview of the existing approaches, they are categorized by notable differences in their architecture or used features. This categorization focuses on the EL aspect of the approaches. 

For each approach, it is mentioned what datasets were used in the corresponding paper. Only a subset of the datasets was directly annotated with Wikidata identifiers. Hence, datasets are mentioned, which do not occur in \Cref{sec:datasets}.

\begin{table*}[tbh!]
    \centering
    \begin{threeparttable}
         \begin{tabularx}{\textwidth}{p{3cm}YYYYYY}
            \toprule
             \textbf{Approach}&\textbf{Labels/\allowbreak Aliases} & \textbf{Descriptions}& \textbf{Knowledge graph structure} & \textbf{Hyper-relational structure} & \textbf{Types} & \textbf{Additional Information} \\ \midrule
            OpenTapioca~\cite{DBLP:journals/corr/abs-1904-09131} & \cmark & \xmark & \cmark & \cmark & \cmark & \xmark  \\
            Falcon 2.0~\cite{DBLP:journals/corr/abs-1912-11270}&\cmark&\xmark&\cmark\tnotex{tn:tabcwc3}&\xmark &\xmark & \xmark\\
            Arjun~\cite{mulang2020encoding}&\cmark&\xmark&\xmark &\xmark& \xmark & \xmark\\
            VCG~\cite{DBLP:conf/starsem/SorokinG18}&\cmark&\xmark& \cmark& \xmark & \xmark & \xmark \\
            KBPearl~\cite{DBLP:journals/pvldb/LinLXLC20} & \cmark & \xmark& \cmark & \xmark & \xmark & \xmark \\
            PNEL~\cite{Banerjee2020} & \cmark & \cmark& \cmark & \xmark & \xmark & \xmark
            \\
            Mulang et al.~\cite{DBLP:journals/corr/abs-2008-05190} & \cmark & \cmark~\tnotex{tn:tabcwc2}& \cmark & \xmark & \xmark & \xmark
            \\
            Perkins~\cite{perkins2020separating} & \cmark & \xmark& \cmark & \xmark & \xmark & \xmark \\
            NED using DL on Graphs~\cite{DBLP:journals/corr/abs-1810-09164}&\cmark&\xmark&\cmark&\xmark & \xmark & \xmark\\
            Huang et al.~\cite{huangentity} & \cmark & \cmark& \cmark & \xmark & \xmark & Wikipedia \\
            Boros et al.~\cite{borosrobust} & \xmark & \xmark & \xmark & \xmark & \cmark & Wikipedia, DBpedia \\
            Provatorov et al.~\cite{Provatorova2020} & \cmark & \cmark & \xmark & \xmark & \xmark & Wikipedia \\ 
            Labusch and Neudecker~\cite{Labusch2020} & \xmark & \xmark & \xmark & \xmark & \xmark & Wikipedia \\
            Botha et al.~\cite{Botha2020} & \xmark & \xmark & \xmark& \xmark & \xmark & Wikipedia \\
            Hedwig~\cite{DBLP:conf/lrec/KlangN20}&\cmark&\cmark& \cmark& \xmark & \xmark & Wikipedia\\
            Tweeki~\cite{tweeki:wnut20} & \cmark & \xmark & \xmark & \xmark & \cmark & Wikipedia \\
            \bottomrule
        \end{tabularx}
        \begin{tablenotes}
            \setlength{\columnsep}{0.8cm}
            \setlength{\multicolsep}{0cm}
            \begin{multicols}{2}
                \small
                \item[2] \label{tn:tabcwc2} Appears in the set of triples used for disambiguation
                \item[1] \label{tn:tabcwc3} Only querying the existence of triples
            \end{multicols}
        \end{tablenotes}
    \end{threeparttable}
    \caption{Comparison between the utilized Wikidata characteristics of each approach.}
    \label{tab:comparison_approaches_wikidata}
\end{table*}

\subsection{Entity Linking} 
\label{subsec:el}
\subsubsection{Language model-based approaches}
The approach by Mulang et al.~\cite{DBLP:journals/corr/abs-2008-05190} is tackling the EL problem with transformer models~\cite{Vaswani2017}. It is assumed that the candidate entities are given. For each entity, the labels of 1-hop and 2-hop triples are extracted. Those are then concatenated together with the utterance and the entity mention. The concatenation is the input of a pre-trained transformer model. With a fully connected layer on top, it is then optimized according to a binary cross-entropy loss. This architecture results in a similarity measure between the entity and the entity mention.
The examined models are the transformer models Roberta~\cite{DBLP:journals/corr/abs-1907-11692}, XLNet~\cite{DBLP:conf/nips/YangDYCSL19} and the DCA-SL model~\cite{DBLP:conf/emnlp/YangGLTZWCHR19}.
The approach was evaluated on three datasets with no focus on certain documents or domains: ISTEX-1000~\cite{DBLP:journals/corr/abs-1904-09131}, Wikidata-Disamb~\cite{DBLP:journals/corr/abs-1810-09164} and AIDA-CoNLL~\cite{hoffart-etal-2011-robust}. AIDA-CoNLL is a popular dataset for evaluating EL but has Wikipedia as the target. ISTEX-1000 focuses on research documents, and Wikidata-Disamb is an open-domain dataset.
There is no global coherence technique applied.
Overall, up to 2-hop triples of any kind are used. For example, labels, aliases, descriptions, or general relations to other entities are all incorporated. It is not mentioned if the hyper-relational structure in the form of qualifiers was used.
On the one hand, the purely language-based EL results in less need for retraining if the KG changes as shown by other approaches~\cite{Botha2020, Wu2019}. This is the case due to the reliance on sub-word embeddings and pre-training via the chosen transformer models. If full word-embeddings were used, the inclusion of new words would make retraining necessary. 
Still, an evaluation of the model on the zero-shot EL task is missing and has to be done in the future.
The reliance on the triple information might be problematic for long-tail entities which are rarely referred to and are part of fewer triples. Nevertheless, a lack of available context information is challenging for any EL approach relying on it.

The approach designed by Botha et al.~\cite{Botha2020} tackles multilingual EL. It is also crosslingual. That means it can link entity mentions to entities in a knowledge graph in a language different from the utterance one. The idea is to train one model to link entities in utterances of 100+ different languages to a KG containing not necessarily textual information in the language of the utterance. While the target KG is Wikidata, they mainly use Wikipedia descriptions as input. This is the case as extensive textual information is not available in Wikidata. 
The approach resembles the Wikification method by Wu et al.~\cite{Wu2019} but extends the training process to be multilingual and targets Wikidata.
Candidate generation is done via a dual-encoder architecture. Here, two BERT-based transformer models~\cite{Devlin2019} encode both the context-sensitive mentions and the entities to the same vector space.
The mentions are encoded using local context, the mention and surrounding words, and global context, the document title. Entities are encoded by using the Wikipedia article description available in different languages. In both cases, the encoded CLS-token are projected to the desired encoding dimension. The goal is to embed mentions and entities in such a way that the embeddings are similar.
The model is trained over Wikipedia by using the anchors in the text as entity mentions. There exists no limitation that the used Wikipedia articles have to be available in all supported languages. If an article is missing in the English Wikipedia but available in the German one, it is still included.
Now, after the model is trained, all entities are embedded. The candidates are generated by embedding the mention and searching for the nearest neighbors.
A cross-encoder is employed to rank the entity candidates, which cross-encodes entity description and mention text together by concatenating and feeding them into a BERT model. Final scores are obtained, and the entity mention is linked.
The model was evaluated on the cross-lingual EL dataset TR2016\textsuperscript{hard}~\cite{Tsai2016} and the multilingual EL dataset Mewsli-9~\cite{Botha2020}. Furthermore, it was tested how well it performs on an English-only dataset called WikiNews-2018~\cite{Gillick2019}.
Wikidata information is only used to gather all the Wikipedia descriptions in the different languages for all entities. 
The approach was tested on zero- and few-shot settings showing that the model can handle an evolving knowledge graph with newly added entities that were never seen before. This is also more easily achievable due to its missing reliance on the graph structure of Wikidata or the structure of Wikipedia. It is the case that some Wikidata entities do not appear in Wikipedia and are therefore invisible to the approach. 
But as the model is trained on descriptions of entities in multiple languages, it has access to many more entities than only the ones available in the English Wikipedia. 

\subsubsection{Language model and graph embeddings-based approaches}
The master thesis by Perkins~\cite{perkins2020separating} is performing candidate generation by using anchor link probability over Wikipedia and locality-sensitive hashing (LSH)~\cite{Gionis1999}  over labels and mention bi-grams. Contextual word embeddings of the utterance (ELMo~\cite{DBLP:conf/naacl/PetersNIGCLZ18}) are used together with KG embeddings (TransE~\cite{DBLP:conf/nips/BordesUGWY13}), calculated over Wikipedia and Wikidata, respectively.
The context embeddings are sent through a recurrent neural network. The output is concatenated with the KG embedding and then fed into a feed-forward neural network resulting in a similarity measure between the KG embedding of the entity candidate and the utterance.
It was evaluated on the AIDA-CoNLL~\cite{hoffart-etal-2011-robust} dataset. 
Wikidata is used in the form of the calculated TransE embeddings. Hyper-relational structures like qualifiers are not mentioned in the thesis and are not considered by the TransE embedding algorithm and, thus, probably not included. 
The used KG embeddings make it necessary to retrain when the Wikidata KG changes as they are not dynamic. 

\subsubsection{Word and graph embeddings-based approaches}
In 2018, Cetoli et al.~\cite{DBLP:journals/corr/abs-1810-09164} evaluated how different types of basic neural networks perform solely over Wikidata. Notably, they compared the different ways to encode the graph context via neural methods, especially the usefulness of including topological information via GNNs~\cite{Sperduti1997, wu2020comprehensive} and RNNs~\cite{Hochreiter1997}. 
There is no candidate generation as it was assumed that the candidates are available.
The process consists of combining text and graph embeddings. The text embedding is calculated by applying a Bi-LSTM over the Glove Embeddings of all words in an utterance. The resulting hidden states are then masked by the position of the entity mention in the text and averaged. A graph embedding is calculated in parallel via different methods utilizing GNNs or RNNs. The end score is the output of one feed-forward layer having the concatenation of the graph and text embedding as its input. It represents if the graph embedding is consistent with the text embedding. 
Wikidata-Disamb30~\cite{DBLP:journals/corr/abs-1810-09164} was used for evaluating the approach. Each example in the dataset also contains an ambiguous negative entity, which is used during training to be robust against ambiguity.
One crucial problem is that those methods only work for a single entity in the text. Thus, it has to be applied multiple times, and there will be no information exchange between the entities.  While the examined algorithms do utilize the underlying graph of Wikidata, the hyper-relational structure is not taken into account. The paper is more concerned with comparing how basic neural networks work on the triples of Wikidata. Due to the pure analytical nature of the paper, the usefulness of the designed approaches to a real-world setting is limited. The reliance on graph embeddings makes it susceptible to change in the Wikidata KG.

\subsection{Entity Recognition and Entity Linking}
\label{subsec:erel}
The following methods all include ER in their EL process. 

\subsubsection{Language model-based approaches}
In connection to the \emph{CLEF 2020 HIPE challenge}~\cite{Ehrmann2020a}, multiple approaches~\cite{borosrobust, Labusch2020, Provatorova2020} for ER and EL of historical newspapers on Wikidata were developed. Documents were available in English, French and German. Three approaches with a focus on the English language are described in the following. Differences in the usage of Wikidata between the languages did not exist. Yet, the approaches were not multilingual as different models were used and/or retraining was necessary for different languages.

Boros et al.~\cite{borosrobust} tackled ER by using a BERT model with a CRF layer on top, which recognizes the entity mentions and classifies the type. During the training, the regular sentences are enriched with misspelled words to make the model robust against noise.
For EL, a knowledge graph is built from Wikipedia, containing Wikipedia titles, page ids, disambiguation pages, redirects and link probabilities between mentions and Wikipedia pages are calculated. The link probability between anchors and Wikipedia pages is used to gather entity candidates for a mention. 
The disambiguation approach follows an already existing method~\cite{Kolitsas2018}. Here, the utterance tokens are embedded via a Bi-LSTM. The token embeddings of a single mention are combined. Then similarity scores between the resulting mention embedding and the entity embeddings of the candidates are calculated. The entity embeddings are computed according to Ganea and Hofmann~\cite{Ganea2017}. These similarity scores are combined with the link probability and long-range context attention, calculated by taking the inner product between an additional context-sensitive mention embedding and an entity candidate embedding. The resulting score is a local ranking measure and is again combined with a global ranking measure considering all other entity mentions in the text.
In the end, additional filtering is applied by comparing the DBpedia types of the entities to the ones classified during the ER. If the type does not match or other inconsistencies apply, the entity candidate gets a lower rank. Here, they also experimented with Wikidata types, but this resulted in a performance decrease.
As can be seen, technically, no Wikidata information besides the unsuccessful type inclusion is used. Thus, the approach resembles more of a Wikification algorithm. Yet, they do link to Wikidata as the HIPE task dictates it, and therefore, the approach was included in the survey. New Wikipedia entity embeddings can be easily added~\cite{Ganea2017} which is an advantage when Wikipedia changes. Also, its robustness against erroneous texts makes it ideal for real-world use. 
This approach reached SOTA performance on the CLEF 2020 HIPE challenge.

Labusch and Neudecker~\cite{Labusch2020} also applied a BERT model for ER. For EL, they used mostly Wikipedia, similar to Boros et al.~\cite{borosrobust}. They built a knowledge graph containing all person, location and organization entities from the German Wikipedia. Then it was converted to an English knowledge graph by mapping from the German Wikipedia Pages via Wikidata to the English ones. This mapping process resulted in the loss of numerous entities. The candidate generation is done by embedding all Wikipedia page titles in an Approximative Nearest Neighbour index via BERT. Using this index, the neighboring entities to the mention embedding are found and used as candidates. For ranking, anchor-contexts of Wikipedia pages are embedded and fed into a classifier together with the embedded mention-context, which outputs whether both belong to the same entity. This is done for each candidate for around 50 different anchor contexts. Then, multiple statistics on those similarity scores and candidates are calculated, which are used in a Random Forest model to compute the final ranks. 
Similar to the previous approach, Wikidata was only used as the target knowledge graph, while information from Wikipedia was used for all the EL work. Thus, no special characteristics of Wikidata were used. The approach is less affected by a change of Wikidata due to similar reasons as the previous approach. This approach lacks performance compared to the state of the art in the HIPE task. The knowledge graph creation process produces a disadvantageous loss of entities, but this might be easily changed.

Provatorov et al.~\cite{Provatorova2020} used an ensemble of fine-tuned BERT models for ER. The ensemble is used to compensate for the noise of the OCR procedure. The candidates were generated by using an ElasticSearch index filled with Wikidata labels. The candidate's final rank is calculated by taking the search score, increasing it if a perfect match applies and finally taking the candidate with the lowest Wikidata identifier number (indicating a high popularity score). They also created three other methods of the EL approach: (1) The ranking was done by calculating cosine similarity between the embedding of the utterance and the embedding of the same utterance with the mention replaced by the Wikidata description. Furthermore, the score is increased by the Levenshtein distance between the entity label and the mention. (2) A variant was used where the candidate generation is enriched with historical spellings of Wikidata entities. (3) The last variant used an existing tool~\cite{Hulst2020}, which included contextual similarity and co-occurrence probabilities of mentions and Wikipedia articles. In the tool, the final disambiguation is based on the ment-norm method by Le and Titov~\cite{Le2018} .
The approach uses Wikidata labels and descriptions in one variant of candidate ranking. Beyond that, no other characteristics specific to Wikidata were considered. Overall, the approach is very basic and uses mostly pre-existing tools to solve the task. The approach is not susceptible to a change of Wikidata as it is mainly based on language and does not need retraining. 

The approach designed by Huang et al.~\cite{huangentity} is specialized in short texts, mainly questions.
The ER is performed via a pre-trained BERT model \cite{Devlin2019} with a single classification layer on top, determining if a token belongs to an entity mention.
The candidate search is done via an ElasticSearch\footnote{\url{https://www.elastic.co/elasticsearch/}} index, comparing the entity mention to labels and aliases by exact match and Levenshtein distance.
The candidate ranking uses three similarity measures to calculate the final rank. A CNN is used to compute a character-based similarity between entity mention and candidate label. This results in a similarity matrix whose entries are calculated by the cosine similarity between each character embedding of both strings. 
The context is included in two ways. First, between the utterance and the entity description, by embedding the tokens of each sequence through a BERT model. Again, a similarity matrix is built by calculating the cosine similarity between each token embedding of both utterance and description. The KG is also considered by including the triples containing the candidate as a subject. For each such triple, a similarity matrix is calculated between the label concatenation of the triple and the utterance. The most representative features are then extracted out of the matrices via max-pooling, concatenated and fed into a two-layer perceptron.
The approach was evaluated on the WebQSP~\cite{DBLP:conf/starsem/SorokinG18} dataset, which is composed of short questions from web search logs.
Wikidata labels, aliases and descriptions are utilized. Additionally, the KG structure is incorporated through the labels of candidate-related triples. This is similar to the approach by Mulang et al.~\cite{DBLP:journals/corr/abs-2008-05190}, but only 1-hop triples are used. There is also no hyper-relational information considered.
Due to its reliance on text alone and using a pre-trained language model with sub-word embeddings, it is less susceptible to changes of Wikidata. While the approach was not empirically evaluated on the zero-shot EL task, other approaches using language models (LM)~\cite{Logeswaran2019,Botha2020, Wu2019} were and indicate a good performance. 

\subsubsection{Word embedding-based approaches}
\emph{Arjun}~\cite{mulang2020encoding} tries to tackle specific challenges of Wikidata like long entity labels and implicit entities. Published in 2020, Arjun is an end-to-end approach utilizing the same model for ER and EL. It is based on an Encoder-Decoder-Attention model. First, the entities are detected via feeding Glove~\cite{Pennington2014} embedded tokens of the utterance into the model and classifying each token as being an entity or not. Afterward, candidates are generated in the same way as in Falcon 2.0~\cite{DBLP:journals/corr/abs-1912-11270} (see \Cref{subsubsec:rule}). The candidates are then ranked by feeding the mention, the entity label, and its aliases into the model and calculating the score. The model resembles a similarity measure between the mention and the entity labels. 
Arjun was trained and evaluated on the T-REx~\cite{elsahar2019t} dataset consisting of extracts out of various Wikipedia articles. 
It does not use any global ranking. Wikidata information is used in the form of labels and aliases in the candidate generation and candidate ranking. The model was trained and evaluated using GloVe embeddings, for which new words are not easily addable. New entities are therefore not easily supported. However, the authors claim that one can replace them with other embeddings like BERT-based ones.  While those proved to perform quite well in zero-shot EL~\cite{Botha2020, Wu2019}, this was usually done with more context information besides labels. Therefore it remains questionable if using those would adapt the approach for zero-shot EL.

\subsubsection{Word and graph embeddings-based approaches}
In 2018, Sorokin and Gurevych~\cite{DBLP:conf/starsem/SorokinG18} were doing joint end-to-end ER and EL on short texts. The algorithm tries to incorporate multiple context embeddings into a mention score, signaling if a word is a mention, and a ranking score, signaling the candidate's correctness. First, it generates several different tokenizations of the same utterance. For each token, a search is conducted over all labels in the KG to gather candidate entities. If the token is a substring of a label, the entity is added.
Each token sequence gets then a score assigned. The scoring is tackled from two sides. On the utterance side, a token-level context embedding and a character-level context embedding (based on the mention) are computed. The calculation is handled via dilated convolutional networks (DCNN) \cite{Yu2016}. On the KG side, one includes the labels of the candidate entity, the labels of relations connected to a candidate entity, the embedding of the candidate entity itself, and embeddings of the entities and relations related to the candidate entity. This is again done by DCNNs and, additionally, by fully connected layers. The best solution is then found by calculating a ranking and mention score for each token for each possible tokenization of the utterance. All those scores are then summed up into a global score. The global assignment with the highest score is then used to select the entity mentions and entity candidates. 
The question-based EL datasets WebQSP~\cite{DBLP:conf/starsem/SorokinG18} and GraphQuestions~\cite{su2016generating} were used for evaluation. GraphQuestions contains multiple paraphrases of the same questions and is used to test the performance on different wordings.
The approach uses the underlying graph, label and alias information of Wikidata. Graph information is used via connected entities and relations. They also use TransE embeddings, and therefore no hyper-relational structure. Due to the usage of static graph embeddings, retraining will be necessary if Wikidata changes.

\emph{PNEL}~\cite{Banerjee2020} is an end-to-end (E2E) model jointly solving ER and EL focused on short texts. PNEL employs a Pointer network~\cite{DBLP:conf/nips/VinyalsFJ15} working on a set of different features. An utterance is tokenized into multiple different combinations. Each token is extended into the (1) token itself, (2) the token and the predecessor, (3) the token and the successor, and (4) the token with both predecessor and successor. For each token combination, candidates are searched for by using the BM25 similarity measure. Fifty candidates are used per tokenization combination. Therefore, 200 candidates (not necessarily 200 distinct candidates) are found per token. For each candidate, features are extracted. Those range from the simple length of a token to the graph embeddings of the candidate entity.
All features are concatenated to a large feature vector. 
Therefore, per token, a sequence of 200 such features vectors exists.
Finally, the concatenation of those sequences of each token in the sentence is then fed into a Pointer network. At each iteration of the Pointer network, it points to one distinct candidate in the network or an \texttt{END} token marking no choice. Pointing is done by computing a softmax distribution and choosing the candidate with the highest probability. Note that the model points to a distinct candidate, but this distinct candidate can occur multiple times. Thus, the model does not necessarily point to only a single candidate of the 200 ones.
PNEL was evaluated on several QA datasets, namely WebQSP~\cite{DBLP:conf/starsem/SorokinG18}, SimpleQuestions~\cite{bordes2015large} and LC-QuAD 2.0~\cite{dubey2019lc}. SimpleQuestions focuses, as the name implies, on simple questions containing only very few entities. LC-QuAD 2.0, on the other hand, contains both, simple and more complex, longer questions including multiple entities.
The entity descriptions, labels and aliases are all used. Additionally, the graph structure is included by TransE graph embeddings, but no hyper-relational information was incorporated.
E2E models can often improve the performance of the ER. Most EL algorithms employed in the industry often use older ER methods decoupled from the EL process. Thus, such an E2E EL approach can be of use. Nevertheless, due to its reliance on static graph embeddings, complete retraining will be necessary if Wikidata changes. 

\subsubsection{Non-NN ML-based approaches}
\emph{OpenTapioca}~\cite{DBLP:journals/corr/abs-1904-09131} is a mainly statistical EL approach published in 2019. 
While the paper never mentions ER, the approach was evaluated with it. In the code, one can see that the ER is done by a SolrTextTagger analyzer of the Solr search platform\footnote{\url{https://lucene.apache.org/solr/}}.  
The candidates are generated by looking up if the mention corresponds to an entity label or alias in Wikidata stored in a Solr collection. Entities are filtered out which do not correspond to the type person, location or organization.
OpenTapioca is based on two main features, which are local compatibility and semantic similarity. First, local compatibility is calculated via a popularity measure and a unigram similarity measure between entity label and mention. The popularity measure is based on the number of sitelinks, PageRank scores, and the number of statements. Second, the semantic similarity strives to include context information in the decision process. All entity candidates are included in a graph and are connected via weighted edges. Those weights are calculated via a statistical similarity measure. This measure includes how likely it is to jump from one entity candidate to another while discounting it by the distance between the corresponding mentions in the utterance. The resulting adjacency matrix is then normalized to a stochastic matrix that defines a Markov Chain. One now propagates the local compatibility using this Markov Chain. Several iterations are then taken, and a final score is inferred via a Support Vector Machine. It supports multiple entities per utterance.
OpenTapioca is evaluated on AIDA-CoNLL~\cite{hoffart-etal-2011-robust}, Microposts 2016~\cite{Rizzo}, ISTEX-1000~\cite{DBLP:journals/corr/abs-1904-09131} and RSS-500~\cite{roder2014n3}. RSS-500 consists of news-based examples and Microposts 2016 focuses on shorter documents like tweets. OpenTapioca was therefore evaluated on many different types of documents.
The approach is only trained on and evaluated for three types of entities: locations, persons, and organizations. 
It facilitates Wikidata-specific labels, aliases, and sitelinks information. More importantly, it also uses qualifiers of statements in the calculation of the PageRank scores. But the qualifiers are only seen as additional edges to the entity.
The usage in special domains is limited due to its restriction to only three types of entities, but this is just an artificial restriction. It is easily updatable if the Wikidata graph changes as no immediate retraining is necessary.

\subsubsection{Graph optimization-based approaches}
\emph{Hedwig}~\cite{DBLP:conf/lrec/KlangN20} is a multilingual entity linker specialized on the TAC 2017~\cite{Ji2017} task but published in 2020. Another entity linker~\cite{DBLP:journals/corr/abs-1903-05498}, developed by the same authors, is not included in this survey as Hedwig is partly an evolution of it. The entities to be linked are limited to only a subset of all possible entity classes. Hedwig employs Wikidata and Wikipedia at the same time. The Entity Recognition uses word2vec embeddings~\cite{Mikolov2013}, character embeddings, and dictionary features where  the character embeddings are calculated via a Bi-LSTM. The dictionary features are class-dependent, but this is not defined in more detail. Those embeddings and features are computed and concatenated for each token.  Afterward, the whole sequence of token features is fed into a Bi-LSTM with a linear chain Conditional Random Field (CRF) layer at the end to recognize the entities. The candidates for each detected entity mention are then generated by using a mention dictionary. The dictionary is created from Wikidata and Wikipedia information, utilizing labels, aliases, titles or anchor texts. The candidates are disambiguated by constructing a graph consisting of all candidate entities, mentions, and occurring words in the utterance. The edges between entities and other entities, words, or mentions have the normalized pointwise mutual information (NPMI) assigned as their weights. The NPMI specifies how frequently two entities, an entity and a mention or an entity and a word, occur together. Those scores are calculated over a Wikipedia dump. Finally, the PageRank of each node in the graph is calculated via power iteration, and the highest-scoring candidates are chosen. 
 
The type classification is used to determine the types of entities, not mentions. As this is only relevant for the TAC 2017 task, the classifier can be ignored. 
The approach was evaluated on the TAC 2017~\cite{Ji2017} dataset, which  focuses on entities of type person, organization, location, geopolitics and facilities. The documents originate from discussion forums and newswire texts.
Labels and aliases from multiple languages are used. It also uses sitelinks to connect the Wikidata identifiers and Wikipedia articles. The paper also claims to use descriptions but does not describe anywhere in what way. No hyper-relational or graph features are used. As it employs class-dependent features, it is limited to the entities of classes specified in the TAC 2017 task. The NPMI weights have to be updated with the addition of new elements in Wikidata and Wikipedia.  

\emph{KBPearl}~\cite{DBLP:journals/pvldb/LinLXLC20}, published in 2020, utilizes EL to populate incomplete KGs using documents. First, a document is preprocessed via Tokenization, POS tagging, NER, noun-phrase chunking, and time tagging. Also, an existing Information Extraction tool is used to extract open triples from the document. They experimented with four different tools (ReVerb~\cite{Fader2011}, MinIE~\cite{Gashteovski2017}, ClausIE~\cite{Corro2013} and Stanford Open IE Tool~\cite{Angeli2015}), Open triples are non-linked triples extracted via an open information extraction tool. The triples consist of a subject, predicate and object in unstructured text. For example, the open triple \texttt{<Ennio Morricone, composed, soundtrack of The Hateful Eight>} can be extracted from "Ennio Morricone, known for numerous famous soundtracks of the Spaghetti Western era, composed the soundtrack of the movie The Hateful Eight.". The triples are processed further by filtering invalid tokens and doing canonicalization. Then, a graph of entities, predicates, noun phrases, and relation phrases is constructed. The candidates are generated by comparing the noun/relation phrases to the labels and aliases of the entities/predicates. The edges between the entities/relations and between entities and relations are weighted by the number of intersecting one-hop statements. The next step is the computation of a maximum dense subgraph. Density is defined by the minimum weighted degree of all nodes \cite{hoffart-etal-2011-robust}. As this problem is NP-hard, a greedy algorithm is used for optimization. New entities relevant for the task of Knowledge Graph Population are identified by thresholding the weighted sum of an entity's incident edges.
Like used here, global coherence can perform sub-optimally since not all entities/relations in a document are related. Thus, two variants of the algorithm are proposed. First, a pipeline version that separates the full document into sentences. Second, a near neighbor mode, limiting the interaction of the nodes in the graph by the distances of the corresponding noun-phrases and relation-phrases.
KBPearl was evaluated on many different datasets: ReVerb38~\cite{DBLP:journals/pvldb/LinLXLC20}, NYT2018~\cite{DBLP:journals/pvldb/LinLXLC20, DBLP:conf/icde/LinC19}, LC-QuAD 2.0~\cite{dubey2019lc}, QALD-7-WIKI~\cite{usbeck20177th}, T-REx~\cite{elsahar2019t}, Knowledge Net~\cite{mesquita2019knowledgenet} and CC-DBP~\cite{glass2018dataset}. These datasets encompass news articles, questions, and general open-domain documents.
The approach includes label and alias information of entities and predicates. Additionally, one-hop statement information is used, but hyper-relational features are not mentioned. However, the paper does not claim that its focus is entirely on Wikidata. Thus, the weak specialization is understandable. While it utilizes EL, the focus of the approach is still on knowledge base population.
No training is necessary, which makes the approach suitable for a dynamic graph like Wikidata.

\subsubsection{Rule-based approaches}
\label{subsubsec:rule}
\emph{Falcon 2.0}~\cite{DBLP:journals/corr/abs-1912-11270} is a fully linguistic approach and a transformation of Falcon 1.0~\cite{DBLP:conf/naacl/SakorMSSV0A19} to Wikidata. Falcon 2.0 was published in 2019, and its focus lies on short texts, especially questions. It links entities and relations jointly. Falcon 2.0 uses entity and relation labels as well as the triples themselves. The relations and entities are recognized by applying linguistic principles. The candidates are then generated by comparing mentions to the labels using the Levenshtein distance. The ranking of the entities and relations is done by creating triples between the relations and entities and checking if the query is successful. The more successful the queries, the higher the candidate will be ranked. If no query is successful, the algorithm returns to the ER phase and splits some of the recognized entities again. As Falcon 2.0 is an extension of Falcon 1.0 from DBpedia to Wikidata, the usage of specific Wikidata characteristics is limited. Falcon 2.0 is tuned for EL on questions and short texts, as well as the English language and it was evaluated on the two QA datasets LC-QuAD 2.0~\cite{dubey2019lc} and SimpleQuestions~\cite{bordes2015large}.
It is not generalizable to longer, more noisy, non-question texts. The used rules follow the structure of short questions. Hence, longer texts consisting of multiple sentences or non-questions are not supported. If the text is grammatically incorrect, the linguistic rules used to parse the utterance would fail. For example, linking Tweets would then be infeasible. 
As it is only based on rules, it is clearly independent of changes in the KG.

\emph{Tweeki}~\cite{tweeki:wnut20} is an approach focusing on unsupervised EL over tweets. The ER is done by a pre-existing Entity Recognizer~\cite{Gardner2018} which also tags the mentions. The candidates are generated by first calculating the link probability between Wikidata aliases over Wikipedia and then searching for the aliases in a dictionary. 
The ranking is done using the link probabilities while pruning all candidates that do not belong to the type provided by the Entity Recognizer. 
Tweeki was evaluated on the accompanied dataset TweekiGold, consisting of random annotated tweets. Additionally, it was tested on the Microposts 2016~\cite{Rizzo} dataset and the datasets by Derczynski~\cite{Derczynski2015} which both also focus on shorter, noisy texts like tweets.
The approach does not need to be trained, making it very suitable for linking entities in tweets. In this document type, often novel entities with minimal context exist.
Regarding features of Wikidata, it uses label, alias and type information. Due to it being unsupervised, changes to the KG do not affect it.

\subsection{Analysis}
\label{sec:approaches_evaluation}

Many approaches include some form of language model or word embedding. This is expected as a large factor of entity linking encompasses the comparison of word-based information. And in that regard, language models like BERT~\cite{Devlin2019} proved very performant in the last years. Furthermore, various language models rely on sub-word or character embeddings which also work on out-of-dictionary words. This is in contrast to regular word-embeddings, which can not cope with words never seen before.
If graph information is part of the approach, the approaches either used graph embeddings, included some coherence score as a feature or created a neighborhood graph on the fly and optimized over it. Some approaches like OpenTapioca, Falcon 2.0 or Tweeki utilized more old-fashioned methods. They either employed classic ML together with some basic features or worked entirely rule-based. 

\subsubsection{Performance}
Table~\ref{tab:dataset_results_el_er} gives an overview of all available results for the approaches performing ER and EL. While results for the EL-only approaches exist, the used measures vary widely. Thus, it is very difficult to compare the approaches. To not withhold the results, they can still be found in the appendix in Table~\ref{tab:dataset_results_el_only} with an accompanying discussion. We aim to fully recover this table and also extend Table~\ref{tab:dataset_results_el_er} in future work.
 
The micro $F_1$ scores are given:
\begin{equation*}
    F_1 = 2 \cdot \frac{p \cdot r}{p +r}
\end{equation*}
where p is the precision $p=\frac{\mathit{tp}}{\mathit{tp} + \mathit{fp}}$ and r is the recall $r=\frac{\mathit{tp}}{\mathit{tp} + \mathit{fn}}$. Here, $\mathit{tp}$ is the number of true positives, $\mathit{fp}$ is the number of false positives and $\mathit{fn}$ is the number of false negatives over a ground truth. Micro $F_1$ means that the scores are calculated over all linked entity mentions and not separately for each document and then averaged. True positives are the correctly linked entity mentions, false positives incorrectly linked entities that do not occur in the set of valid entities and false negatives entities that occur  in the set of valid entities but are not linked to~\cite{Cornolti2013}. 
The approaches were evaluated on many different datasets, which makes comparison very difficult. Additionally, many approaches are evaluated on datasets designed for knowledge graphs different from Wikidata and then mapped. 
Often, the approaches are evaluated on the same dataset but over different subsets, which complicates a comparison even more.
The method by Perkins~\cite{perkins2020separating} was also evaluated on the Kensho Derived Wikimedia Dataset~\cite{KenshoRD}, but it was only used to compare different variants of the designed approach and focused on different amounts of training data. Thus, inclusion in the evaluation table is not reasonable. 

\begin{table*}
    \centering
    \rotatebox{-90}{
    \resizebox{0.95\textheight}{!}{
    \begin{threeparttable}
        \begin{tabular}{ccccccccccccc}
            \toprule
            &\rotatebox[origin=c]{-66}{OpenTapioca~\cite{DBLP:journals/corr/abs-1904-09131}} & 
            \rotatebox[origin=c]{-66}{Falcon 2.0~\cite{DBLP:journals/corr/abs-1912-11270}} &
            \rotatebox[origin=c]{-66}{Arjun~\cite{mulang2020encoding}} & 
            \rotatebox[origin=c]{-66}{VCG~\cite{DBLP:conf/starsem/SorokinG18}} &
            \rotatebox[origin=c]{-66}{KBPearl~\cite{DBLP:journals/pvldb/LinLXLC20}~\tnotex{tn:erel2}} &
            \rotatebox[origin=c]{-66}{PNEL~\cite{Banerjee2020}}&
            \rotatebox[origin=c]{-66}{Huang et al.~\cite{huangentity}} &
            \rotatebox[origin=c]{-66}{Boros et al.~\cite{borosrobust}} &
            \rotatebox[origin=c]{-66}{Provatorov et al.~\cite{Provatorova2020}} &
            \rotatebox[origin=c]{-66}{\parbox{2cm}{Labusch \& Neudecker~\cite{Labusch2020}}} &
            \rotatebox[origin=c]{-66}{Hedwig~\cite{DBLP:conf/lrec/KlangN20}} &
            \rotatebox[origin=c]{-66}{Tweeki~\cite{tweeki:wnut20}}\\
            \toprule 
            AIDA-CoNLL~\cite{hoffart-etal-2011-robust} &0.482~\cite{DBLP:journals/corr/abs-1904-09131}&-&-&-&-&-&-&-&-&-&-&-\\
            Microposts 2016~\cite{Rizzo}&0.087~\cite{DBLP:journals/corr/abs-1904-09131}, 0.148~\cite{tweeki:wnut20}&-&-&-&-&-&-&-&-&-&-&0.248~\cite{tweeki:wnut20} \\
            ISTEX-1000~\cite{DBLP:journals/corr/abs-1904-09131}&0.87~\cite{DBLP:journals/corr/abs-1904-09131}&-&-&-&-&-&-&-&-&-&-&- \\
            RSS-500~\cite{roder2014n3} & 0.335~\cite{DBLP:journals/corr/abs-1904-09131}&-&-&-&-&-&-&-&-&-&-&-\\
            LC-QuAD 2.0~\cite{dubey2019lc}&0.301~\cite{Banerjee2020} & 0.445~\cite{Banerjee2020}&-&
            0.47~\cite{Banerjee2020}&-&0.589~\cite{Banerjee2020}~\tnotex{tn:erel3}&-&-&-&-&-&-
            \\
            LC-QuAD 2.0~\cite{dubey2019lc}~\tnotex{tn:erel8}&0.25~\cite{DBLP:journals/corr/abs-1912-11270} & 0.68~\cite{DBLP:journals/corr/abs-1912-11270}&-&
            &-&-&-&-&-&-&-&-
            \\
            LC-QuAD 2.0~\cite{dubey2019lc}~\tnotex{tn:erel1}&-& 0.320~\cite{Banerjee2020}&-&-&-&0.629~\cite{Banerjee2020}~\tnotex{tn:erel3}&-&-&-&-&-&-
            \\
            Simple-Question&0.20~\cite{Banerjee2020}& 0.41~\cite{Banerjee2020}&-&-&-&0.68~\cite{Banerjee2020}~\tnotex{tn:erel4}&-&-&-&-&-&-\\
            Simple-Question~\cite{bordes2015large}~\tnotex{tn:erel9}&-& 0.63~\cite{DBLP:journals/corr/abs-1912-11270}&-&-&-&-&-&-&-&-&-&-\\
            T-REx~\cite{elsahar2019t}& 0.579~\cite{mulang2020encoding}&-&0.713~\cite{mulang2020encoding}&-&-&-&-&-&-&-&-&-\\
            T-REx~\cite{elsahar2019t}~\tnotex{tn:erel7}& -&-&-&-&0.421~\cite{DBLP:journals/pvldb/LinLXLC20}&-&-&-&-&-&-&-\\
            WebQSP~\cite{yih2016value}&-&-&0.730~\cite{Banerjee2020,DBLP:conf/starsem/SorokinG18}&-&-&0.712~\cite{Banerjee2020}~\tnotex{tn:erel5}&0.780~\cite{huangentity}&-&-&-&-&-\\
            CLEF HIPE 2020~\cite{Ehrmann2020}&-&-&-&-&-&-&-&0.531~\cite{Ehrmann2020a}~\tnotex{tn:erel6}&0.300~\cite{Ehrmann2020a}~\tnotex{tn:erel6}&0.141~\cite{Ehrmann2020a}~\tnotex{tn:erel6}&-&-\\
            TAC2017~\cite{Ji2017}&-&-&-&-&-&-&-&-&-&-&0.582~\cite{DBLP:conf/lrec/KlangN20}\\
            Graph-Questions~\cite{su2016generating}&-&-&0.442~\cite{DBLP:conf/starsem/SorokinG18}&-&-&-&-&-&-&-&-&-\\
            QALD-7-WIKI~\cite{usbeck20177th}&-&-&-&-&0.679~\cite{DBLP:journals/pvldb/LinLXLC20}&-&-&-&-&-&-&-\\
            NYT2018~\cite{DBLP:journals/pvldb/LinLXLC20, DBLP:conf/icde/LinC19}&-&-&-&-&0.575~\cite{DBLP:journals/pvldb/LinLXLC20}&-&-&-&-&-&-&-\\
            ReVerb38~\cite{DBLP:journals/pvldb/LinLXLC20}&-&-&-&-&0.653~\cite{DBLP:journals/pvldb/LinLXLC20}&-&-&-&-&-&-&-\\
            Knowledge Net~\cite{mesquita2019knowledgenet}&-&-&-&-&0.384~\cite{DBLP:journals/pvldb/LinLXLC20}&-&-&-&-&-&-&-\\
            CC-DBP~\cite{glass2018dataset}&-&-&-&-&0.499~\cite{DBLP:journals/pvldb/LinLXLC20}&-&-&-&-&-&-&-\\
            TweekiGold~\cite{tweeki:wnut20}&0.291~\cite{tweeki:wnut20}&-&-&-&-&-&-&-&-&-&-&0.65~\cite{tweeki:wnut20}\\
            Derczynski~\cite{Derczynski2015}&0.14~\cite{tweeki:wnut20}&-&-&-&-&-&-&-&-&-&-&0.371~\cite{tweeki:wnut20}\\
            \bottomrule
        \end{tabular}
        \begin{tablenotes}
            \setlength{\columnsep}{0.8cm}
            \setlength{\multicolsep}{0cm}
            \begin{multicols}{3}
                \small
                \item[1] \label{tn:erel2} NN model
                \item[2] \label{tn:erel3} L model
                \item[3] \label{tn:erel8} 1000 sampled questions from LC-QuAD 2.0
                \item[4] \label{tn:erel1} LC-QuAD 2.0 test set used in KBPearl paper
                \item[5] \label{tn:erel4} S model
                \item[6] \label{tn:erel9} Probably evaluated on train and test set
                \item[7] \label{tn:erel7} Evaluation on subset of T-REx data different to the subset used in Arjun paper
                \item[8] \label{tn:erel5} W model
                \item[9] \label{tn:erel6} Strict mention matching
            \end{multicols}
        \end{tablenotes}
    \end{threeparttable}
    }}
    \caption{Results: ER + EL.}
    \label{tab:dataset_results_el_er}
\end{table*}

Inferring the utility of a Wikidata characteristic from the different approaches' $F_1$-measures is inconclusive due to the sparsity of results.
For ER + EL approaches, most results were available for LC-QuAD 2.0. Yet, no conclusion can be drawn as many approaches were evaluated on different subsets of the dataset. Falcon 2.0 performs well, but it does not substantially rely on Wikidata characteristics. The performance is good as it is designed for simple questions that follow its rules very closely. Arjun performs well on T-REx by mainly using label information, but the number of methods tested on the T-REx dataset is too low to be conclusive. Besides that, PNEL and the approach by Huang et al. also achieve good results; both include a broader scope of Wikidata information in the form of labels, descriptions and graph structure. As HIPE challenge approaches are using Wikidata only marginally and the difference in performance depends more on the robustness against the OCR-introduced noise, comparing them is not providing information on the relevance of Wikidata characteristics.

\subsubsection{Utilization of Wikidata characteristics}
\begin{figure}[hbt!]
    \centering
    \includegraphics[width=\linewidth]{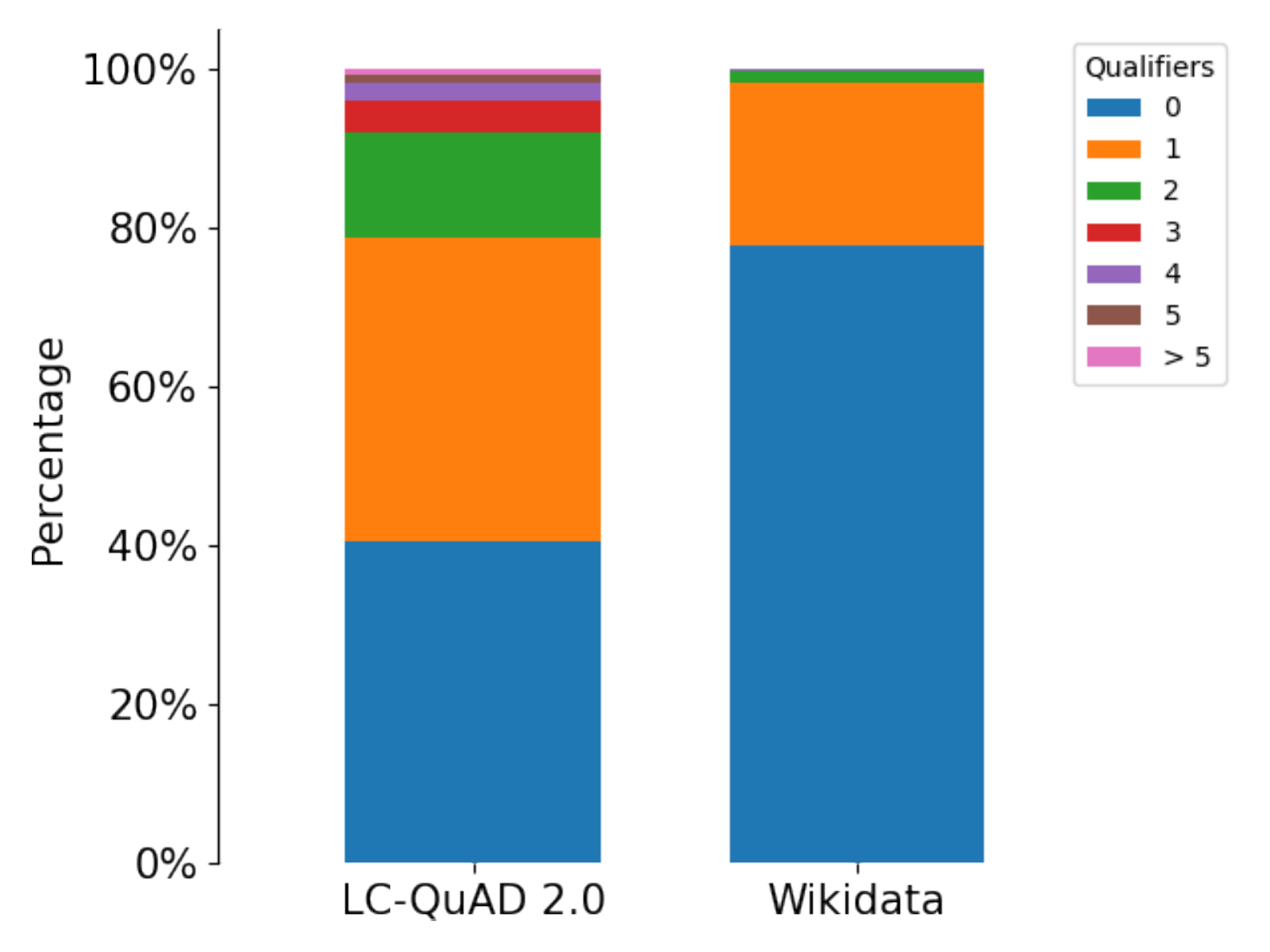}
    \caption{Percentage of statements having the specified number of qualifiers for all LC-QuAD 2.0 and Wikidata entities.}
    \label{fig:qualifiers_bars}
\end{figure}

While some algorithms~\cite{mulang2020encoding} do try to examine the challenges of Wikidata, like more noisy long entity labels, many fail to use most of the advantages of Wikidata's characteristics. If the approaches are using even more information than just the labels of entities and relations, they mostly only include simple n-hop triple information. Hyper-relational information like qualifiers is only used by OpenTapioca but still in a simple manner. This is surprising, as they can provide valuable additional information. As one can see in Figure~\ref{fig:qualifiers_bars}, around half of the statements on entities occurring in the LC-QuAD 2.0 dataset have one or more qualifiers. These percentages differ from the ones in all of Wikidata, but when entities are considered, appearing in realistic use cases like QA, qualifiers are much more abundant.   Thus, dismissing the qualifier information might be critical. The inclusion of hyper-relational graph embeddings could improve the performance of many approaches already using non-hyper-relational ones.
Rank information of statements might be useful to consider, but choosing the best one will probably often suffice. 

Of all approaches, only two algorithms~\cite{huangentity, Banerjee2020} use descriptions explicitly. Others incorporate them through triples too, but more on the side~\cite{DBLP:journals/corr/abs-2008-05190}. Descriptions can provide valuable context information and many items do have them; see Figure~\ref{fig:wikidata_descriptions}. Hedwig~\cite{DBLP:conf/lrec/KlangN20} claims to use descriptions but fails to describe how.  

Two approaches~\cite{DBLP:conf/lrec/KlangN20, Botha2020} demonstrated the usefulness of the inherent multilingualism of Wikidata, notably in combination with Wikipedia. 

\begin{table}[htb!]
    \centering
    \begin{tabularx}{\linewidth}{Xcc}
        \toprule
         \textbf{Approach} & \textbf{Code} & \textbf{Web API} \\ \midrule
         OpenTapioca~\cite{DBLP:journals/corr/abs-1904-09131} & \cmark & \cmark \\
        Falcon 2.0~\cite{DBLP:journals/corr/abs-1912-11270}&\cmark&\cmark\\
        Arjun~\cite{mulang2020encoding}&\cmark&\xmark\\
        VCG~\cite{DBLP:conf/starsem/SorokinG18}&\cmark&\xmark \\
        KBPearl~\cite{DBLP:journals/pvldb/LinLXLC20} & \xmark & \xmark \\
        PNEL~\cite{Banerjee2020} & \cmark & \xmark \\
        Mulang et al.~\cite{DBLP:journals/corr/abs-2008-05190} & \cmark & \xmark\\
        Perkins~\cite{perkins2020separating} & \xmark & \xmark\\
        NED using DL on Graphs~\cite{DBLP:journals/corr/abs-1810-09164}&\cmark&\xmark\\
        Huang et al.~\cite{huangentity} & \xmark & \xmark\\ 
        Boros et al.~\cite{borosrobust} & \xmark & \xmark \\
        Provatorov et al.~\cite{Provatorova2020} & \xmark & \xmark \\
        Labusch and Neudecker~\cite{Labusch2020} & \cmark & \xmark \\
        Botha et al.~\cite{Botha2020} & \xmark & \xmark \\
        Hedwig~\cite{DBLP:conf/lrec/KlangN20}&\xmark&\xmark \\
        Tweeki~\cite{tweeki:wnut20} & \xmark & \xmark\\
        \bottomrule
    \end{tabularx}
    \caption{Availability of approaches.}
    \label{tab:availability}
\end{table}

As Wikidata is always changing, approaches robust against change are preferred. A reliance on transductive graph embeddings~\cite{DBLP:conf/starsem/SorokinG18,Banerjee2020, DBLP:journals/corr/abs-1810-09164, perkins2020separating}, which need to have all entities available during training, makes repeated training necessary. Alternatively, the used embeddings would need to be replaced with graph embeddings, which are efficiently updatable or inductive~\cite{Wang2019, Wang2019a, Teru2020,Baek2020,Albooyeh2020,Hamaguchi2017,Wu2019a, galkin2021nodepiece, Daruna2021}.
The rule-based approach Falcon 2.0~\cite{DBLP:journals/corr/abs-1912-11270} is not affected by a developing knowledge graph but only usable for correctly-stated questions. 
Methods only working on text information~\cite{mulang2020encoding, Provatorova2020, huangentity, Botha2020, DBLP:journals/corr/abs-2008-05190} like labels, descriptions or aliases do not need to be updated if Wikidata changes, only if the text type or the language itself does. This is demonstrated by the approach by Botha et al.~\cite{Botha2020} and the Wikification EL BLINK~\cite{Wu2019}, which mainly use the BERT model and are able to link to entities never seen during training. If word-embeddings instead of sub-word embeddings are used, for example, GloVe~\cite{Pennington2014} or word2vec~\cite{Mikolov2013}, this advantage diminishes as new never-seen labels could not be interpreted. Nevertheless, the ability to support totally unseen new entities was only demonstrated for the approach by Botha et al~\cite{Botha2020}. The other approaches still need to be evaluated on the zero-shot EL task to be certain. 
For approaches~\cite{DBLP:conf/lrec/KlangN20, huangentity, tweeki:wnut20} that rely on statistics over Wikipedia, new entities in Wikidata may sometimes not exist in Wikipedia to a satisfying degree.
As a consequence, only a subset of all entities in Wikidata is supported. This also applies to the approaches by Boros et al.~\cite{borosrobust}, and Labusch and Neudecker~\cite{Labusch2020} which are mostly using Wikipedia information. Additionally, they are susceptible to changes in Wikipedia, especially specific statistics calculated over Wikipedia pages which have to be updated any time a new entity is added.
Botha et al.~\cite{Botha2020} also mainly depend on Wikipedia and thus on the availability of the desired Wikidata entities in Wikipedia itself. Since the approach uses Wikipedia articles in multiple languages, it encompasses many more entities than the previous approaches that focus on Wikipedia. Botha et al.'s~\cite{Botha2020} approach was designed for the zero- and few-shot setting, it is quite robust against changes in the underlying knowledge graph.

\begin{table*}[hb!]
    \centering
    \begin{tabularx}{0.99\linewidth}{lcccc}
        \toprule
         \textbf{Survey} & \textbf{\# Approaches} & \textbf{\# Wikidata Approaches} & \textbf{\# Datasets} & \textbf{\# Wikidata Datasets}    \\ \midrule
         Sevgili et al.~\cite{DBLP:journals/corr/abs-2006-00575} & 30 & 0 & 9 & 0 \\
         Al-Moslmi et al.~\cite{al2020named} & 39 & 0 & 17 & 0\\ 
         Oliveira et al.~\cite{oliveira2020towards}  & 36 & 0 & 32 & 0 \\
         This survey & 16 & 16 & 21 & 11 \\
         \bottomrule
    \end{tabularx}
    \caption{Survey Comparison}
    \label{tab:survey_comparison}
\end{table*}

Approaches relying on statistics~\cite{DBLP:journals/pvldb/LinLXLC20, DBLP:journals/corr/abs-1904-09131} need to update them regularly, but this might be efficiently doable.
Overall, the robustness against change might be negatively affected by static/transductive graph embeddings.

\begin{highlightbox}{\hyperlink{rq3}{Research Question 3}}{How do current Entity Linking approaches exploit the specific characteristics of Wikidata?}
The preceding summary and evaluation of the existing Wikidata Entity Linkers, together with \Cref{tab:comparison_approaches_wikidata} and the descriptions in \Cref{subsec:el,subsec:erel}, provide an overview of all approaches with a focus on the incorporated Wikidata-characteristics.
\end{highlightbox}
\begin{highlightbox}{\hyperlink{rq4}{Research Question 4}}{Which Wikidata characteristics are unexploited by existing Entity Linking approaches?}
The most unexploited characteristics are the descriptions, the hyper-relational structure and the type information, as can be seen in \Cref{tab:comparison_approaches_wikidata}.
Nearly none of the found approaches exploited hyper-relational information in the form of qualifiers. And the one (i.e. OpenTapioca) using them did that in a simple way. As it is confirmed by the benchmarks that the inclusion of those can improve the performance of link prediction~\cite{Galkin2020}, this might also be the case for the task of EL. 
Furthermore, description information is still greatly underutilized. It can be a valuable piece of context information of an entity. Of course, it is not ideal as often the description can be short, especially for long-tail entities. 
A possible way to circumvent this challenge is the recent development of \textbf{Abstract Wikipedia
}~\cite{DBLP:journals/cacm/Vrandecic21}, which will support the multilingual generation of descriptions in the future. 
While some approaches utilize type information, most use them to limit the set of valid entity candidates to instances of only a small subset of all types, namely person, organization and location. This is surprising as the non-included paper by Raiman and Raiman~\cite{DBLP:conf/aaai/RaimanR18} shows that a fine-grained type system can heavily improve the entity linking performance.
As mentioned in \Cref{sec:wikidata}, rank information might also be used by not only including statements of the best rank but also of others. For example, in the case that statements exist which were valid at different points of time, including all could prove useful when linking documents of different ages. But such a special use case is not considered by any existing Wikidata EL approach.
For some more ideas on how to include those characteristics in the future, please refer to \cref{subsec:future}.
\end{highlightbox}

\subsection{Reproducibility}
Not all approaches are available as a Web API or even as source code. An overview can be found in Table~\ref{tab:availability}. 
The number of approaches for Wikidata having an accessible Web API is meager. While the code for some methods exists, this is the case for only half of them. The effort to set up different approaches also varies significantly due to missing instructions or data. 
Thus, we refrained from evaluating and filling the missing results for all the datasets in Tables~\ref{tab:dataset_results_el_only} and~\ref{tab:dataset_results_el_er}. However, we seek to extend both tables in future work.

\section{Related work}
\label{sec:related-work}

While there are multiple recent surveys on EL, none of those are specialized in analyzing EL on Wikidata.

The extensive survey by Sevgili et al.~\cite{DBLP:journals/corr/abs-2006-00575} is giving an overview of all neural approaches from 2015 to 2020. It compares 30 different approaches on nine different datasets. 
According to our criteria, none of the included approaches focuses on Wikidata. The survey also discusses the current state of the art of domain-independent and multi-lingual neural EL approaches. However, the influence of the underlying KG was not of concern to the authors. It is not described in detail how they found the considered approaches.

In the survey by Al-Moslmi et al.~\cite{al2020named}, the focus lies on ER and EL approaches over KGs in general. It considers approaches from 2014 to 2019.  It gives an overview of the different approaches of ER, Entity Disambiguation, and EL. A distinction between Entity Disambiguation and EL is made, while our survey sees Entity Disambiguation as a part of EL. The roles of different domains, text types, or languages are discussed.  The authors considered 89 different approaches and tools. 
Most approaches were designed for DBpedia or Wikipedia, some for Freebase or YAGO, and some to be KG-agnostic. Again, none focused on Wikidata. $F_1$ scores were gathered on 17 different datasets.
Fifteen algorithms, for which an implementation or a WebAPI was available, were evaluated using GERBIL~\cite{Roeder2018}.

Another survey~\cite{oliveira2020towards} examines recent approaches, which employ holistic strategies. Holism in the context of EL is defined as the usage of domain-specific inputs and metadata, joint ER-EL approaches, and collective disambiguation methods. Thirty-six research articles were found which had any holistic aspect - none of the designed approaches linked explicitly to Wikidata.

A comparison of the number of approaches and datasets included in the different surveys can be found in Table~\ref{tab:survey_comparison}. 

If we go further into the past, the existing surveys~\cite{shen2014entity, DBLP:journals/tacl/LingSW15} are not considering Wikidata at all or only in a small amount as it is still a rather recent KG in comparison to the other established ones like DBpedia, Freebase or YAGO. For an overview of different KGs on the web, we refer the interested reader to the paper by Heist et al.~\cite{DBLP:series/ssw/HeistHRP20}.

No found survey focused on the differences of EL over different knowledge graphs, respectively, on the particularities of EL over Wikidata. 
\section{Discussion} \label{sec:discussion}
\subsection{Current Approaches, Datasets and their Drawbacks}
\paragraph{Approaches.}
The number of algorithms using Wikidata is small; the number of algorithms using Wikidata solely is even smaller. Most algorithms employ labels and alias information contained in Wikidata. Some deep learning-based algorithms leverage the underlying graph structure, but the inclusion of that information is often superficial. The same information is also available in other KGs. Additional statement-specific information like qualifiers is used by only one algorithm (OpenTapioca), and even then, it only interprets qualifiers as extra edges to the item. Thus, there is no inclusion of the actual structure of a hyper-relation. Information like the descriptions of items that are providing valuable context information is also rarely used. Wikidata includes type information, but almost none of the existing algorithms utilize it to do more than to filter out entities that are not desired to link in general. An exception is perhaps Tweeki, though it only uses types during ER.

It seems that most of the authors developed approaches for Wikidata due to it being popular and up-to-date while not specifically utilizing its structure. With small adjustments, many would also work on any other KG. 
Besides the less-dedicated utilization of specific characteristics of Wikidata, it is also notable that there is no clear focus on one of the essential characteristics of Wikidata, continual growth. Many approaches use static graph embeddings, which need to be retrained if the KG changes. EL algorithms working on Wikidata, which are not usable on future versions, seem unintuitive. But there also exist some approaches which can handle change. They often rely on more extensive textual information, which is again challenging due to the limited amount of such data in Wikidata. Wikidata descriptions do exist, but only short paragraphs are provided, in general, insufficient to train a language model. To compensate, Wikipedia is included, which provides this textual information. It seems like Wikidata as the target KG with its language-agnostic identifiers and the easily connectable Wikipedia with its multilingual textual information are a great pair. But surprisingly, most methods do use either Wikipedia or Wikidata. A combination happens rarely but seems very fruitful, as can be seen via the performance of the multilingual EL by Botha et al.~\cite{Botha2020}. Though even this approach still uses Wikidata only sparsely.  

None of the investigated approaches' authors tried to examine the performance between different versions of Wikidata. Since continuous evolution is a central characteristic of Wikidata, a temporal analysis would be reasonable. 
As we are confronted with a fast-growing ocean of knowledge, taking into account the change of Wikidata and hence developing approaches that are robust against that change will undoubtedly be useful for numerous applications and their users. 

This survey aimed to identify the extent to which the current state of the art in Wikidata EL is utilizing the characteristics of Wikidata. As only a few are using more information than on other established KGs, there is still much potential for future research. 

\paragraph{Datasets.} Only a limited number of datasets were created entirely with Wikidata in mind exist. Many datasets used are still only mapped versions of datasets created for other knowledge graphs. Multilingualism is present so far that some datasets contain documents in different languages. However, only different documents for different languages are available. Having the same documents in multiple languages would be more helpful for an evaluation of multilingual Entity Linkers.  The fact that the Wikidata is ever-changing is also not genuinely considered in any datasets. Always providing the dump version on which the dataset was created is advisable. A big advantage for the community  is that datasets from very different domains like news, forums, research, tweets exist. The utterances can also vary from shorter texts with only a few entities to large documents with many entities.
The difficulty of the datasets significantly differs in the ambiguity of the entity mentions. 
The datasets also differ in quality. Some were automatically created and others annotated manually by experts.
There are no unanimously agreed-upon datasets used for Wikidata EL.
Of course, a single dataset can not exist as different domains and text types make different approaches, and hence datasets necessary.

\subsection{Future Research Avenues}
\label{subsec:future}

In general, Wikidata EL could be improved by including the following aspects:

\paragraph{Hyper-relational statements.}
The qualifier and rank information of Wikidata could be suitable to do EL on time-sensitive utterances~\cite{DBLP:conf/acl/AgarwalSCHW18}. The problem revolves around utterances that talk about entities from different time points and spans and thus, the referred entity can significantly diverge.
The usefulness of other characteristics of Wikidata, e.g., references, may be limited but could make EL more challenging due to the inclusion of contradictory information. Therefore, research into the consequences and solutions of conflicting information would be advisable.
Another possibility would be to directly include the qualifier information via the KG embeddings. For example, the StarE~\cite{Galkin2020} embedding includes qualifiers directly during training. It performs superior over regular embeddings on the task of link prediction if enough statements have qualifiers assigned. This is promising but whether this directly applies to EL approaches, which use such embeddings, has to be evaluated.

\paragraph{More extensive type information.}
While type information is incorporated by some linkers, it is generally done to simply limit the candidate space to the three main types: location, organization and person. But Raiman and Raiman~\cite{DBLP:conf/aaai/RaimanR18} showed that a more extensive system of types proves very effective on the task of EL. If an adequate typing system is chosen and the correct type of an entity mention is available, an entity linker can achieve a near-perfect performance. Especially as Wikidata has a much more fine-grained and noisy type system than other KGs, evaluating the performance of entity linkers, which incorporate types, is of interest. 
While most approaches use types directly to limit the candidate space, incorporating them indirectly via type-aware ~\cite{Zhang2020} or hierarchy-sensitive embeddings~\cite{Chami2020, Nayyeri2021, DBLP:conf/nips/BalazevicAH19} might also prove useful for EL.
But note that the incorporation of type information heavily depends on the performance of the type classifier, and the difficulty of the type classification task again depends on the type system. Nevertheless, an improved type classification would directly benefit type-utilizing entity linkers. 

\paragraph{Inductive or efficiently trainable knowledge graph embeddings.}
To reiterate, due to the fast rate of change of Wikidata, approaches are necessary, which are more robust to such a dynamic KG. Continuously retraining transductive embeddings is intractable, so more sophisticated methods like inductive or efficiently retrievable graph embeddings are a necessity~\cite{Wang2019, Wang2019a, Teru2020,Baek2020,Albooyeh2020,Hamaguchi2017,Wu2019a, galkin2021nodepiece, Daruna2021}.
For example, the embedding by Albooyeh et al.~\cite{Albooyeh2020} can be employed, which can handle out-of-sample entities. These are entities, which were not available at training time, but are connected to entities, which were existing.
To go even further, NodePiece~\cite{galkin2021nodepiece}, the KG-embedding counterpart of sub-word embeddings like BERT, works by relying on only a small subset of anchor nodes and all relations in the KG. While it uses a fraction of all nodes, it still is able to achieve performance competitive with transductive embeddings on the task of link prediction. By being independent of most nodes in a KG, one can include new entities (in the form of nodes) without having to retrain.  As an alternative, standard continual learning approaches could be employed to learn new data while being robust against catastrophic forgetting. An examination of the performance of popular techniques in the context of KG embeddings can be found in the paper by Daruna et al.~\cite{Daruna2021}.

\paragraph{Item label and description information in multiple languages for multilingual EL.}
Multilingual or cross-lingual EL is already tackled with Wikidata but currently mainly by depending on Wikipedia. Using the available multilingual label/description information in a structured form together with the rich textual information in Wikipedia could move the field forward. The approach by Botha et al.~\cite{Botha2020}, which could be seen as an extension of BLINK~\cite{Wu2019}, performs very well on the task of cross- and multilingual EL. 
For example, the approach by Mulang et al.~\cite{DBLP:journals/corr/abs-2008-05190}, which fully relies on label information, could be extended in a similar way as BLINK was extended. Instead of only using labels (of items and properties) in the English language, training the model directly in multiple languages could prove effective. Additionally, multilingual description information might be used too. We are convinced that also investigations into the linking of long-tail entities are needed.

It seems like there exist no commonly agreed-on Wikidata EL datasets, as shown by a large number of different datasets the approaches were tested on. Such datasets should try to represent the challenges of Wikidata like the time-variance, contradictory triple information, noisy labels, and multilingualism.

\section*{Acknowledgments}
We acknowledge the support of the EU project TAILOR (GA 952215), the Federal Ministry for Economic Affairs and Energy (BMWi) project SPEAKER (FKZ 01MK20011A), the German Federal Ministry of Education and Research (BMBF) projects and excellence clusters ML2R (FKZ 01 15 18038 A/B/C), MLwin (01S18050 D/F), ScaDS.AI (01/S18026A) as well as the Fraunhofer Zukunftsstiftung project JOSEPH. The authors also acknowledge the financial support by the Federal Ministry for Economic Affairs and Energy of Germany in the project CoyPu (project number 01MK21007G).

\begin{appendix}
\section{KG-agnostic Entity Linkers}
\label{sec:appendix}

AGDISTIS~\cite{DBLP:conf/ecai/UsbeckNRGCAB14} is an EL approach expecting already marked entity mentions. It expects a KG dump available in the Turtle format~\cite{beckett2014rdf}. For candidate generation, first, an index is created which contains all available entities and their labels. They are extracted from the available Turtle dump. The input entity mention is first normalized by reducing plural and genitive forms and removing common affixes. Furthermore, if an entity mention consists of a substring of a preceding entity mention, the succeeding one is directly mapped to the preceding one. Additionally, the space of possible candidates can be limited by configuration. Usually, the candidate space is reduced to organizations, persons and locations. The candidates are then searched for over the index by comparing the reduced entity mention with the labels in the index using trigram similarity. No candidates are included, which contain time information inside the label.
After gathering all candidates of all entity mentions in the utterance, the candidates are ranked by building a temporary graph. Starting with the candidates as the initial nodes, the graph is expanded breadth-first by adding the adjacent nodes and the edges in-between. It is done to some previously set depth. This results in a partly connected graph containing all candidates. 
Then the HITS-algorithm~\cite{Kleinberg1999} is run and the most authoritative candidate nodes are chosen per entity mention. Thus, the approach is performing a global entity coherence optimization. 
The approach uses label and alias information for building the index. Type information can be used to restrict the candidate space and the KG structure is utilized during the candidate ranking. 

\paragraph{MAG.} MAG~\cite{Moussallem2017} is a multilingual extension of AGDISTIS. Again, no ER is performed. The same label index as used in AGDISTIS is employed. Besides that, the following additional indices were created: 
\begin{itemize}
    \item A person index, containing the person names and the variations in different languages
    \item A rare references index containing textual descriptions of entities
    \item An acronym index based on the commercial STANDS4~\footnote{\label{footnote1}\url{http://www.abbreviations.com/}} data
    \item A context index containing semantic embeddings of Concise Bounded Description~\footnote{\url{https://www.w3.org/Submission/CBD/}}
\end{itemize} 
During candidate generation, it is first checked if the entity mention corresponds to an acronym. If it is one, no further preprocessing is done. If not, the entity mention is normalized by removing special characters, changing the casing and splitting camel-cased words. After preprocessing, the candidates are searched by first checking for exact matches, then searching via trigram similarity. If this still did not produce any candidates, the entity mention is stemmed and the search is repeated. If a mention is an acronym, the candidate list is expanded with the corresponding entities.
Then, more candidates are searched by taking an entity mention and the set of all entity mentions in the utterance. Those are used to build a tf-idf search query over the context index. All returned candidates are then first filtered by trigram similarity between entity mention and candidate. A second filtering is applied by counting the number of direct connections between the remaining candidates and the candidates of the other entity mentions. The candidates with too few links are pruned away. All the candidates of the entity mention are then sorted by their popularity (calculated via PageRank~\cite{page1999pagerank}) and the top 100 are returned.
Then, the entities are disambiguated in nearly the same way as done by AGDISTIS. The only difference is the option to use PageRank instead of HITS to rank the final candidates. 
Additionally to the properties already used by AGDISTIS, item descriptions are incorporated via the context index.

DoSeR~\cite{DBLP:conf/esws/ZwicklbauerSG16} also expects already marked entity mentions. The linker focuses being to link to multiple knowledge graphs simultaneously. Here, they support RDF-based KGs and entity-annotated document (EAD) KGs (e.g., Wikipedia). The KGs are split into core and optional KGs. Core KGs contain the entities to which one wants to link. Optional KGs complement the core KGs with additional data.
First, an index is created which includes the entities of all core KGs. In the index, the labels or surface forms, a semantic embedding, and each entity's prior popularity are stored. 
The semantic embeddings are computed by using Word2Vec. For EAD-KGs, the different documents are taken and all words, which are not pointing to entities, are removed. All remaining words are replaced with the corresponding entity identifier. These sequences are then used to train the embeddings. For RDF-KGs, a Random Walk is performed over the graph and the resulting sequences are used to train the embeddings. The succeeding node is chosen with a probability corresponding to the reciprocal of the number of edges it got. The same probability is used to sometimes jump to another arbitrary node in the graph. 
The prior probability is calculated by either using the number of incoming/outgoing edges in the RDF-KG or the number of annotations that point to the entity in the EAD KG. If type information is available, the entity space can be limited here too. 
First, candidates are generated by searching for exact matches and then the AGDISTIS candidate generation is used to find more candidates. 
The candidates are disambiguated, similar to the way AGDISTIS and MAG are doing it. First, a graph is built though not a complete graph but a $K$-partite graph where $K$ is the number of all entity mentions. Edges exist only between candidates of different entities. Using the complete graph resulted in a loss of performance. After the graph is created, PageRank is done to score the different entities coherently. The edge weights correspond to the (normalized) cosine similarity of the semantic embeddings of the two connected entities. Additionally, at any point during PageRank computation, it is possible to jump to an arbitrary node with a certain probability. This probability depends on the prior popularity of the entity.
It uses label information, the knowledge graph structure and type information (if desired). 
\begin{table*}[tbh!]
    \centering
     \begin{tabularx}{\textwidth}{p{3cm}YYYYYY}
        \toprule
         \textbf{Approach}&\textbf{Labels/\allowbreak Aliases} & \textbf{Descriptions}& \textbf{Knowledge graph structure} & \textbf{Hyper-relational structure} & \textbf{Types} & \textbf{Additional Information} \\ \midrule
         AGDISTIS~\cite{DBLP:conf/ecai/UsbeckNRGCAB14} & \cmark & \xmark & \cmark & \xmark & \cmark & \\
         MAG~\cite{Moussallem2017} & \cmark & \cmark & \cmark & \xmark & \cmark & STANDS4~\ref{footnote1} \\
         DoSeR~\cite{DBLP:conf/esws/ZwicklbauerSG16} & \cmark & \xmark & \cmark & \xmark & \cmark & \\
        \bottomrule
    \end{tabularx}
    \caption{Comparison between the utilized Wikidata characteristics of each KG-agnostic approach.}
    \label{tab:comparison_KG_agnostic_approaches_wikidata}
\end{table*}



\section{EL-only results and discussion}

\begin{table*}
    \centering
    \begin{threeparttable}
        \begin{tabular}{cccP{3cm}c}
                \toprule
                &\rotatebox[origin=c]{-66}{Mulang et al.~\cite{DBLP:journals/corr/abs-2008-05190}}
                &\rotatebox[origin=c]{-66}{LSH-ELMo model~\cite{perkins2020separating}}
                &\rotatebox[origin=c]{-66}{\parbox{3cm}{NED using DL on Graphs~\cite{DBLP:journals/corr/abs-1810-09164}~\tnotex{tn:elonly1}}} &\rotatebox[origin=c]{-66}{Botha et al.~\cite{Botha2020}}\\
                \midrule
                AIDA-CoNLL~\cite{hoffart-etal-2011-robust}&0.9494~\cite{DBLP:journals/corr/abs-2008-05190}~\tnotex{tn:elonly6}~\tnote{,}~\tnotex{tn:elonly3}&0.73~\cite{perkins2020separating}&-&-\\
                ISTEX-1000~\cite{DBLP:journals/corr/abs-1904-09131}&0.9261~\cite{DBLP:journals/corr/abs-2008-05190}~\tnotex{tn:elonly4}&-&-&-\\
                Wikidata-Disamb~\cite{DBLP:journals/corr/abs-1810-09164}&0.9235~\cite{DBLP:journals/corr/abs-2008-05190}~\tnotex{tn:elonly5}&-&0.916~\cite{DBLP:journals/corr/abs-1810-09164}&-\\
                Mewsli-9~\cite{Botha2020}  & - & - & - & 0.91~\cite{Botha2020}~\tnotex{tn:elonly7} \\
                \bottomrule
        \end{tabular}
        \begin{tablenotes}
            \setlength{\columnsep}{0.8cm}
            \setlength{\multicolsep}{0cm}
            \begin{multicols}{3}
                \small
                \item[1] \label{tn:elonly1} Model with best result
                \item[2] \label{tn:elonly6} Accuracy instead of $F_1$
                \item[3] \label{tn:elonly3} DCA-SL used
                \item[4] \label{tn:elonly4} XLNet used
                \item[5] \label{tn:elonly5} Roberta used
                \item[6] \label{tn:elonly7} Recall instead of $F_1$
            \end{multicols}
        \end{tablenotes}
    \end{threeparttable}
    \caption{Results: EL-only.}
    \label{tab:dataset_results_el_only}
\end{table*}

The results for EL-only approaches can be found in Table~\ref{tab:dataset_results_el_only}.
AIDA-CoNLL results are available for three of the four approaches, but the results for one is the accuracy instead of the $F_1$-measures. 
The available labels for each item and property make language-model-based approaches possible that perform quite well~\cite{DBLP:journals/corr/abs-2008-05190}. No approaches are available to compare to the one by Botha et al.~\cite{Botha2020}, but the result demonstrates the promising performance of multilingual EL with Wikidata as the target KG.

\begin{table*}[]
    \centering
    \begin{tabular}{ll}
    \toprule
        Dataset & Links \\
         \midrule
         T-REx~\cite{elsahar2019t} & \url{https://hadyelsahar.github.io/t-rex} \\ 
         NYT2018~\cite{DBLP:journals/pvldb/LinLXLC20, DBLP:conf/icde/LinC19} & not found \\
         ISTEX-1000~\cite{DBLP:journals/corr/abs-1904-09131} & \url{https://github.com/wetneb/opentapioca/blob/master/data} \\
         LC-QuAD 2.0~\cite{dubey2019lc} &  \url{https://github.com/AskNowQA/LC-QuAD2.0/tree/master/dataset}\\
         Knowledge Net~\cite{mesquita2019knowledgenet} & \url{https://github.com/diffbot/knowledge-net/tree/master/dataset}
         \\
         KORE50DYWC~\cite{noullet2020kore} & \url{https://www.aifb.kit.edu/web/KORE_50%5EDYWC} \\
         Kensho Derived Wikimedia Dataset~\cite{KenshoRD} &  \url{https://www.kaggle.com/kenshoresearch/kensho-derived-wikimedia-data} \\
         CLEF HIPE 2020~\cite{Ehrmann2020} & \url{https://github.com/impresso/CLEF-HIPE-2020/tree/master/data}   \\
         Mewsli-9~\cite{Botha2020} & \url{https://metatext.io/datasets/mewsli-9-}  \\
         TweekiData~\cite{tweeki:wnut20} & \url{https://github.com/ucinlp/tweeki/tree/main/data/Tweeki_data} \\
         TweekiGold~\cite{tweeki:wnut20}& \url{https://github.com/ucinlp/tweeki/tree/main/data/Tweeki_gold}\\
         \bottomrule
    \end{tabular}
    \caption{Links to datasets}
    \label{tab:dataset_links}
\end{table*}

\end{appendix}
\FloatBarrier
\nocite{label}
\bibliography{collection}

\end{document}